\documentclass[lettersize,journal]{IEEEtran}

\IEEEoverridecommandlockouts                              

\usepackage{graphicx}
\usepackage{cite}
\usepackage{amsmath}
\usepackage{adjustbox}
\usepackage{array}
\usepackage{multicol}
\usepackage{float}
\usepackage{setspace}
\usepackage{bm}
\usepackage{amsfonts}
\usepackage{amsthm}
\usepackage{makecell}
\usepackage{multirow}
\usepackage[hidelinks]{hyperref}
\usepackage{bm}
\usepackage{color}
\allowdisplaybreaks
\usepackage{graphicx}
\usepackage[caption=false,font=footnotesize]{subfig}

\newtheorem{thm}{Theorem}

\newtheorem{rem}{Remark}
\newtheorem{cor}{Corollary}
\newtheorem{prop}{Proposition}

\newtheorem{definition}[thm]{Definition}

\title{Robust Lateral String Stability and Error Attenuation
for Heterogeneous Vehicle Platoons}

\author{Sixu Li,~\IEEEmembership{Student Member,~IEEE}, Swaroop Darbha,~\IEEEmembership{Fellow,~IEEE}, Yang Zhou,~\IEEEmembership{Member,~IEEE}
\thanks{(Corresponding authors: Swaroop Darbha, Yang Zhou)}
\thanks{The authors are with Texas A$\&$M University, College Station, TX 77843, USA (e-mail: sixuli@tamu.edu; dswaroop@tamu.edu; yangzhou295@tamu.edu).}}%

\begin{document}
\maketitle
\thispagestyle{empty}
\pagestyle{empty}

\begin{abstract}
Keeping an entire vehicle platoon close to its planned path is essential for navigating road boundaries and obstacles safely. Followers may rely on predecessor information for lateral guidance, which couples their lateral dynamics and allows errors to propagate downstream. This paper develops a robust lateral string stability framework to study this propagation and achieve downstream error attenuation in heterogeneous platoons under varying longitudinal motion and local perturbations. Arc-length-based feedback linearization gives each vehicle the same nominal lateral error dynamics, with parameter uncertainty and external disturbances represented as local perturbations. We define robust $\mathcal L_2$ lateral string stability in terms of platoon-length-uniform error bounds and derive a sufficient condition for achieving it. Furthermore, we show that the same condition guarantees platoon-length-uniform $\mathcal L_\infty$ error bounds. For the considered control formulations, predecessor
tracking (PT) can accumulate lateral errors, whereas
desired-path tracking (DT) prevents this accumulation.
We propose a learn-from-predecessor (LFP) strategy
with explicit gains for downstream error attenuation
and characterize how intervehicle perturbation
differences limit this attenuation. Simulations validate the theoretical findings under vehicle heterogeneity, varying longitudinal motion, and local perturbations.
\end{abstract}


\section{Introduction}
\IEEEPARstart{C}{onnected} and automated vehicle platoons offer
opportunities to improve traffic throughput and energy efficiency
\cite{vegamoor2019review}. Realizing these benefits requires
attention to how errors arising at individual vehicles propagate
through the platoon, as their amplification can compromise
platoon performance. This concern motivates string stability,
which has been studied extensively for longitudinal spacing
errors \cite{swaroop1994string,swaroop2002string}, with controller
design and information exchange playing important roles
\cite{swaroop1999constant,seiler2004disturbance,zheng2015stability}.

\begin{figure}[h]
    \centering
    \subfloat[Safe leader avoidance]{%
        \includegraphics[width=0.48\columnwidth]{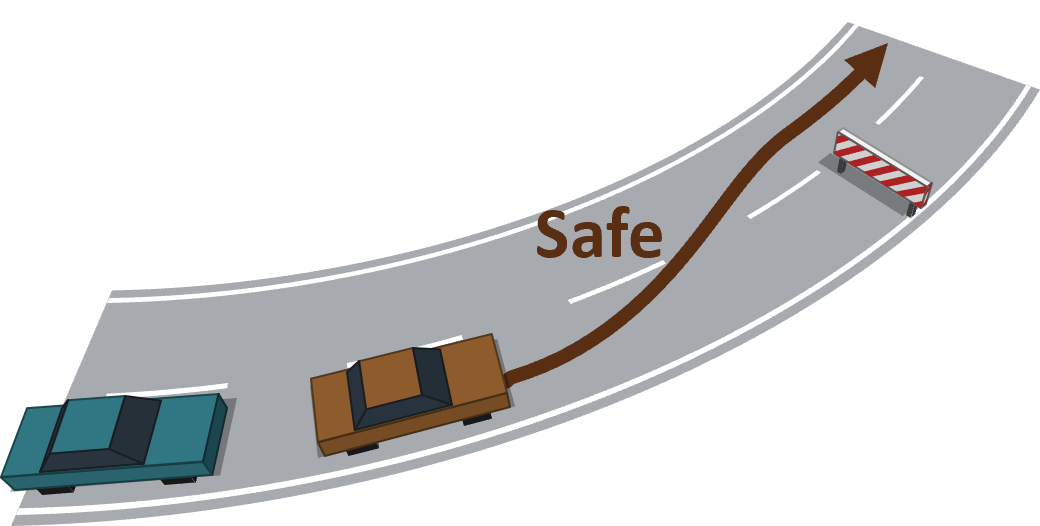}%
        \label{fig:occlusion_leader}%
    }
    \hfil
    \subfloat[Delayed follower response]{%
        \includegraphics[width=0.48\columnwidth]{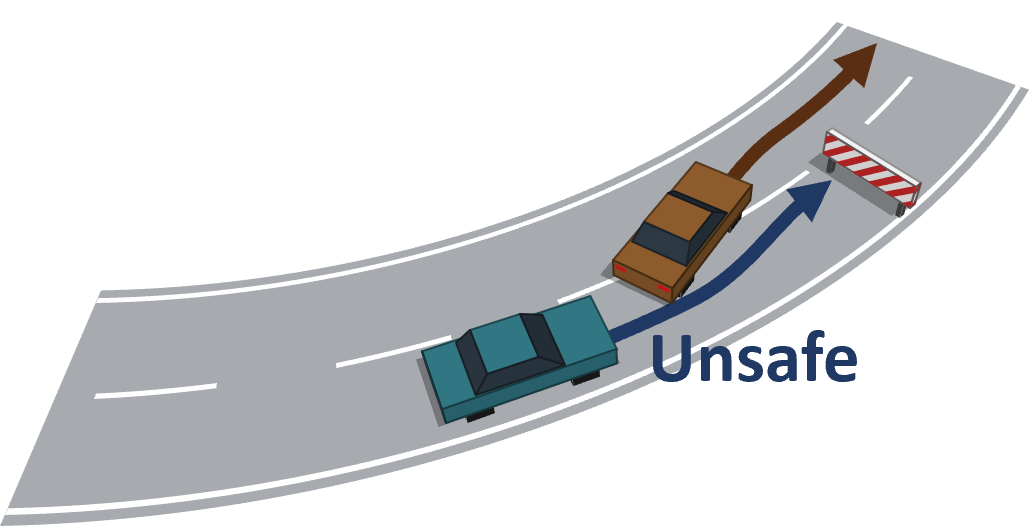}%
        \label{fig:occlusion_follower}%
    }
    \caption{Sensor occlusion in a vehicle platoon. }
    \label{fig:occlusion_scenario}
\end{figure}

Lateral coordination is also important when platoons follow curved roads or perform obstacle-avoidance maneuvers. In a platoon, a vehicle can obstruct the field of view of onboard cameras and LiDARs on the vehicle behind it \cite{zhang2021safe,yu2019occlusion}. Consider the obstacle-avoidance scenario in Fig.~\ref{fig:occlusion_scenario}. The lead vehicle detects the obstacle and plans a safe avoidance path, as shown in Fig.~\ref{fig:occlusion_scenario}(a), but blocks the second vehicle's view of the obstacle. By the time the second vehicle detects it, the remaining distance may be insufficient to independently plan and execute a safe maneuver, as shown in Fig.~\ref{fig:occlusion_scenario}(b). This delayed perception effect can further accumulate downstream as the platoon size increases. This motivates using the leader's planned path as the desired path for the platoon, allowing followers to benefit from the leader's earlier perception and planning \cite{hassanain2020string,liu2020robust,li2026lateral}. Guidance for following this path may be obtained through onboard sensing of a predecessor's traveled path or through vehicle-to-vehicle (V2V) communication of the desired path. When predecessor information
is used for lateral control, the vehicles' lateral
dynamics become interconnected, allowing lateral
errors to propagate downstream. This motivates
examining whether these errors amplify along the
platoon through the study of lateral string stability
\cite{li2026lateral}.

The spatial nature of lateral motion also matters when
assessing error propagation. Road curvature and physical
boundaries vary along the path, while vehicles occupy
different locations at the same time instant. Comparing
their lateral errors at that instant therefore mixes the
effects of different road geometries with error propagation
along the platoon. Comparing errors as vehicles pass the
same desired-path location avoids this ambiguity. This
motivates the arc-length-based framework for lateral string
stability developed in \cite{li2026lateral} for homogeneous
platoons at constant longitudinal speed.

Practical platoons, however, may contain vehicles
with different physical parameters and operate under
varying longitudinal motion. Moreover, parameter
uncertainty and external disturbances introduce local
perturbations that affect lateral error propagation.
This paper develops a robust lateral string stability
framework for this setting and proposes a
learn-from-predecessor (LFP) strategy for downstream
error attenuation. The analysis establishes conditions
for bounded lateral errors throughout the platoon
and characterizes how differences between vehicles'
local perturbations affect the guaranteed attenuation.

\subsection{Related Work and Motivation}
Longitudinal string stability was formally established in the work of Swaroop and Hedrick \cite{swaroop1994string,swaroop2002string} to characterize how spacing errors propagate along automated vehicle platoons and has since been studied under various spacing policies and information topologies \cite{swaroop1999constant,seiler2004disturbance,zheng2015stability}. Subsequent work has addressed communication imperfections, including packet loss, noise, and delay \cite{vegamoor2021string,ma2024selection,ma2025robust}, and extended the analysis to mixed traffic and human-driven vehicles \cite{jin2014dynamics,jin2016optimal,tian2025physically}. Theoretical developments include $\mathcal{L}_p$-based input--output formulations \cite{ploeg2013lp} and disturbance string stability accounting for disturbances at individual vehicles \cite{besselink2017string}. Among the norm-based criteria, $\mathcal{L}_2$ string stability is widely used because error amplification in stable LTI interconnections can be assessed in the frequency domain \cite{seiler2004disturbance,ploeg2013lp}. Under additional assumptions, it also yields bounds on peak errors \cite{vegamoor2021string}.

Lateral string stability has received comparatively less
attention, with existing studies adopting different error
measures and stability criteria. For example, \cite{hassanain2020string} employed
$\mathcal{H}_\infty$ norms of transfer functions to characterize
lateral string stability, while \cite{liu2020robust} considered
$\mathcal{L}_\infty$ attenuation of tracking error vectors and
identified limitations of relying solely on predecessor
information. Bounded tracking errors and the benefits of
leader information were examined in
\cite{somisetty2024lateral}. However, much of this early work
focused on straight-line paths. Moreover, the quantities
analyzed for lateral string stability in these studies,
such as yaw rate or predecessor-relative errors, do not
directly characterize lateral deviations from the desired
path, which determine whether vehicles maintain the clearance
from obstacles and road boundaries intended by the path
planner. These limitations motivated the arc-length-based
framework in \cite{li2026lateral}, which explicitly considers
desired-path errors and compares them at common locations
along curved paths. That framework introduced an
$\mathcal{L}_2$ lateral string stability definition,
established a necessary and sufficient condition for the
considered stable LTI interconnections, and examined the
effects of information modes, control strategies, and error
measures. Its analysis, however, was restricted to homogeneous
platoons at constant longitudinal speed.

Practical platoons may consist of vehicles with different physical parameters and operate under varying longitudinal speeds and accelerations, motivating an extension beyond the homogeneous, constant-speed setting. Moreover, parameter uncertainty and external disturbances can affect lateral control performance and error propagation. These considerations motivate developing a robust notion of lateral string stability and studying downstream error attenuation for heterogeneous platoons under varying longitudinal motion.

\subsection{Proposed Framework and Contributions}

This paper develops a framework for robust lateral string stability and downstream error attenuation in heterogeneous vehicle platoons under varying longitudinal motion and local perturbations. The main contributions are as follows:
\begin{enumerate}
    \item A feedback linearization formulation is developed directly in the arc-length domain to account for vehicle heterogeneity and varying longitudinal motion. It yields the same nominal lateral error dynamics for each vehicle, with parameter uncertainty and external disturbances represented as local perturbations, enabling a unified analysis of lateral error propagation.

    \item A robust $\mathcal{L}_2$ lateral string stability notion is introduced to characterize platoon-length-uniform lateral error bounds in terms of initial conditions and local perturbations. A sufficient condition is established for the considered one-predecessor LTI interconnections. Furthermore, the same condition guarantees platoon-length-uniform $\mathcal{L}_\infty$ lateral error bounds.

    \item Predecessor tracking (PT) and learn-from-predecessor (LFP) control are formulated for the heterogeneous, perturbed setting. The analysis establishes that the considered PT configuration fails to guarantee robust lateral string stability and derives sufficient learning-gain conditions under which LFP satisfies it.

    \item An explicit LFP gain design makes the predecessor-to-follower propagation map contractive in both the $\mathcal{L}_2$ and $\mathcal{L}_\infty$ norms. Under zero initial conditions, exact downstream error attenuation is established for common perturbations, while error envelopes characterize how intervehicle perturbation differences limit the guaranteed attenuation.
\end{enumerate}

Numerical studies of the reduced arc-length error dynamics
first validate the theoretical bounds and examine how
perturbation repeatability affects downstream error attenuation.
Simulations with heterogeneous vehicle dynamics then evaluate
performance under varying longitudinal motion, parameter
uncertainty, and road-bank disturbances. Both studies compare
PT, desired-path tracking (DT) without learning, and LFP to
distinguish the benefits of desired-path information from the
additional error attenuation achieved through LFP.

\subsection{Organization of the Paper}
The remainder of this paper is organized as follows. Section~\ref{sec:arc_length_fl_heterogeneous} presents the arc-length-based feedback linearization for heterogeneous vehicles and the local perturbation model. Section~\ref{sec:platoon_control} formulates platoon control under the two information modes. Section~\ref{sec:robust_lateral_string_stability} introduces robust lateral string stability and establishes sufficient conditions. Section~\ref{sec:configuration_analysis} analyzes robust lateral string stability under the two information modes and establishes downstream error attenuation results for LFP. Section~\ref{sec:numerical_evaluation} presents the numerical evaluation, and Section~\ref{sec:chapter4_summary} concludes the paper.

\section{Arc-Length-Based Feedback Linearization for Heterogeneous Vehicle Platoons}
\label{sec:arc_length_fl_heterogeneous}

This section develops an arc-length-based feedback linearization
formulation for heterogeneous vehicle platoons under varying
longitudinal motion. The formulation gives each vehicle the same
nominal lateral error dynamics for the subsequent platoon-level
analysis. Parameter uncertainty and external disturbances are then
represented as local perturbations.

Consider a platoon of $N$ vehicles, indexed by $i\in\{1,\ldots,N\}$,
following a common desired path. Each vehicle uses a reference path
for lateral control, which need not coincide with the desired path.
The reference-path selection and intervehicle information exchange
are specified in Section~\ref{sec:platoon_control}. The physical
parameters and longitudinal motion may differ among vehicles.

\subsection{Time-domain tracking model}
\label{subsec:heterogeneous_time_model}

We adopt the tracking error model developed in \cite{ch3preprint},
which expresses the bicycle dynamics relative to a reference path
while explicitly retaining varying longitudinal speed and
acceleration. Its final form is recalled here with vehicle-specific
parameters; the detailed derivation is given in \cite{ch3preprint}.

For vehicle $i$, let $m_i$ and $I_{z,i}$ denote the mass and yaw
moment of inertia, $C_{f,i}$ and $C_{r,i}$ the front and rear tire
cornering stiffnesses, and $a_i$ and $b_i$ the distances from the
center of gravity to the front and rear axles. Let $v_{x,i}$,
$\theta_i$, and $\delta_{f,i}$ denote the longitudinal speed,
vehicle heading, and front steering angle, respectively. A dot denotes differentiation with respect to time.
For the subsequent arc-length parameterization, longitudinal
speed is assumed bounded away from zero:
$v_{x,i}(t)\ge v_{\min}>0$.

\begin{figure}[h]
    \centering
    \includegraphics[width=0.42\columnwidth]{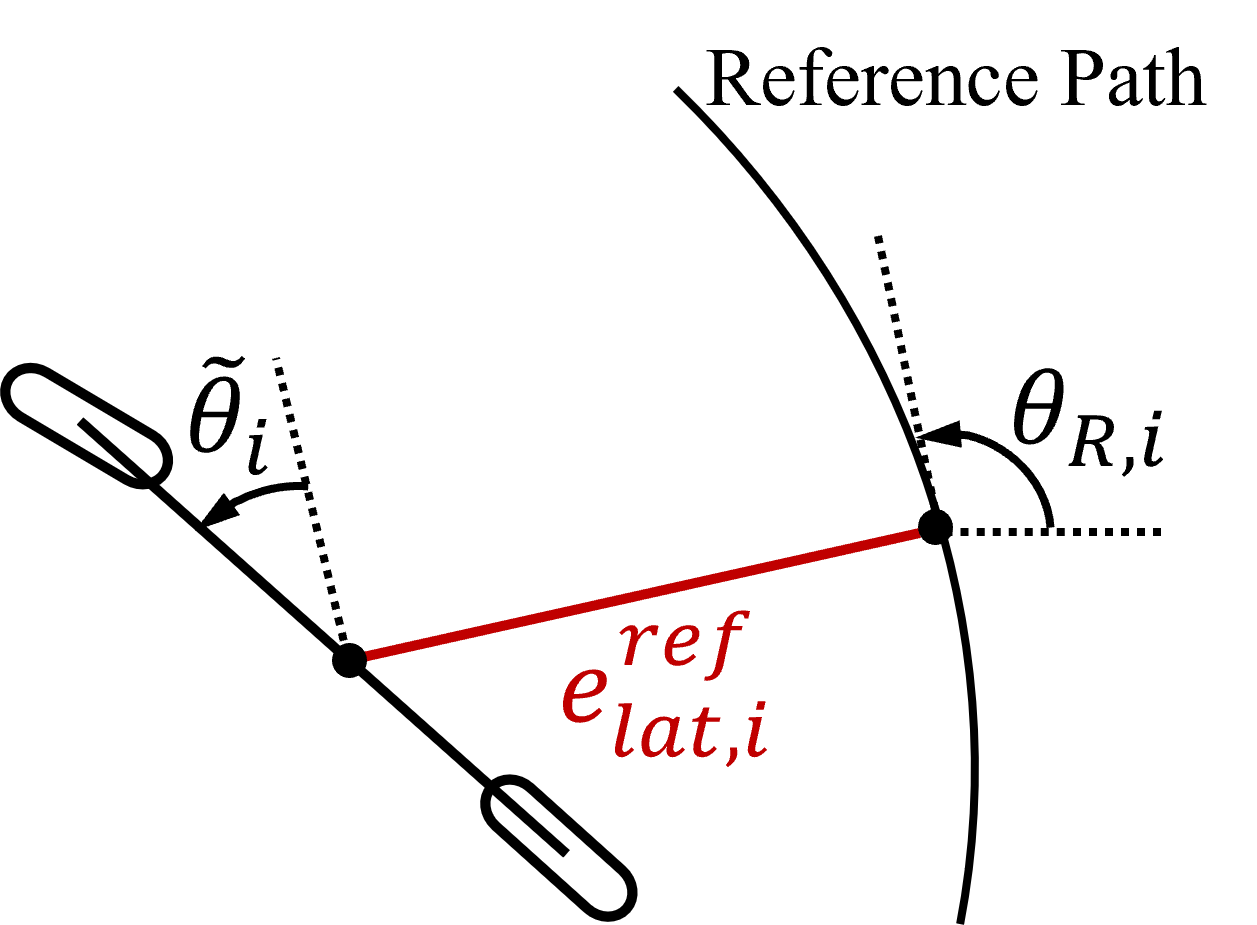}
    \caption{Lateral and heading errors relative to the reference path.}
    \label{fig:hetero_error_geometry}
\end{figure}

As illustrated in Fig.~\ref{fig:hetero_error_geometry},
$e_{lat,i}^{ref}$ is the signed normal displacement of the
vehicle's center of gravity from its reference path.
The reference-path heading at the projection point is
$\theta_{R,i}$, and the heading error is
$\tilde\theta_i:=\theta_i-\theta_{R,i}$.

 With these definitions, the
tracking error dynamics are
{\small
\begin{align}
\label{eq:hetero_time_elat}
\ddot e_{lat,i}^{ref}
={}&
\underbrace{
\begin{aligned}
&\left(\tfrac{C_{f,i}+C_{r,i}}{m_i}
+\dot v_{x,i}\right)\tilde\theta_i-\tfrac{C_{f,i}+C_{r,i}}{m_i v_{x,i}}
\dot e_{lat,i}^{ref}
-\tfrac{a_iC_{f,i}-b_iC_{r,i}}{m_i v_{x,i}}
\dot{\tilde\theta}_i
\end{aligned}
}_{f_{1,i}}
\notag\\
&+\underbrace{\tfrac{C_{f,i}}{m_i}}_{g_{1,i}}\delta_{f,i}
+\underbrace{
\left(-v_{x,i}
-\tfrac{a_iC_{f,i}-b_iC_{r,i}}{m_i v_{x,i}}\right)
}_{d_{1,i}}\dot\theta_{R,i},
\\
\label{eq:hetero_time_heading}
\ddot{\tilde\theta}_i
={}&
\underbrace{
\begin{aligned}
&\tfrac{a_iC_{f,i}-b_iC_{r,i}}{I_{z,i}}\tilde\theta_i-\tfrac{a_iC_{f,i}-b_iC_{r,i}}{I_{z,i}v_{x,i}}
\dot e_{lat,i}^{ref}
-\tfrac{a_i^2C_{f,i}+b_i^2C_{r,i}}{I_{z,i}v_{x,i}}
\dot{\tilde\theta}_i
\end{aligned}
}_{f_{2,i}}
\notag\\
&+\underbrace{\tfrac{a_iC_{f,i}}{I_{z,i}}}_{g_{2,i}}\delta_{f,i}
+\underbrace{
\left(-\tfrac{a_i^2C_{f,i}+b_i^2C_{r,i}
+I_{z,i}\dot v_{x,i}}{I_{z,i}v_{x,i}}\right)
}_{d_{2,i}}\dot\theta_{R,i}.
\end{align}
}
The $\dot v_{x,i}$ terms are retained because longitudinal speed
is not assumed constant.

\subsection{Arc-length-based reparameterization}
\label{subsec:arc_length_objective}

Lateral string stability is assessed by comparing vehicle errors at
the same location along the desired path \cite{li2026lateral}.
Let $l_d$ denote the desired-path arc length, and let $t_i(l_d)$
denote the time at which the desired-path projection of vehicle $i$
reaches location $l_d$. For any time-domain signal $q_i(t)$, define
\begin{equation}
\label{eq:arc_evaluation}
    q_i(l_d):=q_i\bigl(t_i(l_d)\bigr).
\end{equation}
Thus, different vehicles' signals at the same $l_d$ are evaluated
at their respective traversal times.

For example,
$e_{lat,i}^{ref}(l_d)=e_{lat,i}^{ref}(t_i(l_d))$ and
$\dot v_{x,i}(l_d)=\dot v_{x,i}(t_i(l_d))$.
When a dotted quantity is written as a function of $l_d$, the dot
continues to denote differentiation with respect to time; only its
value is reparameterized. A prime denotes differentiation with
respect to $l_d$.

Following the arc-length formulation in \cite{li2026lateral},
the desired-path arc length and time are related approximately by
\begin{equation}
\label{eq:ld_dot}
    \frac{d l_d}{dt}\approx v_{x,i}(l_d).
\end{equation}
Together with the positive-speed assumption, this provides a
monotone spatial coordinate. Applying the chain rule gives
\begin{align}
\label{eq:arc_first_derivative}
    \dot e_{lat,i}^{ref}(l_d)
    &=\frac{d e_{lat,i}^{ref}(l_d)}{d l_d}
      \frac{d l_d}{dt} \approx v_{x,i}(l_d)
    \left(e_{lat,i}^{ref}(l_d)\right)'.
\end{align}
Differentiating once more with respect to time yields
\begin{align}
\label{eq:arc_second_derivative}
    \ddot e_{lat,i}^{ref}(l_d)
    \approx{}&
    \dot v_{x,i}(l_d)
    \left(e_{lat,i}^{ref}(l_d)\right)'
    +v_{x,i}^{2}(l_d)
    \left(e_{lat,i}^{ref}(l_d)\right)''.
\end{align}

Substituting \eqref{eq:arc_second_derivative} into
\eqref{eq:hetero_time_elat} yields the arc-length-based
lateral error dynamics
\begin{align}
\label{eq:arc_length_io}
    \left(e_{lat,i}^{ref}(l_d)\right)''
    =&\frac{1}{v_{x,i}^{2}(l_d)}
    \Big[
      f_{1,i}(l_d)+g_{1,i}\delta_{f,i}(l_d)
      \notag\\
      &~~+d_{1,i}(l_d)\dot\theta_{R,i}(l_d)
      -\dot v_{x,i}(l_d)\left(e_{lat,i}^{ref}(l_d)\right)'
    \Big].
\end{align}
In \eqref{eq:arc_length_io}, each time-domain quantity is
evaluated at $t_i(l_d)$, when vehicle $i$'s projection onto
the desired path reaches $l_d$. This spatial model provides
the basis for the feedback linearization developed next.

\subsection{Arc-length-based feedback linearization}
\label{subsec:arc_fl}
The coefficients in \eqref{eq:arc_length_io} depend on each
vehicle's physical parameters and longitudinal motion.
To obtain the same nominal lateral error dynamics across
vehicles, we use feedback linearization to impose
\begin{equation}
\label{eq:generic_arc_dynamics}
    \left(e_{lat,i}^{ref}(l_d)\right)''=\mu_i(l_d),
\end{equation}
where $\mu_i$ is a virtual control input. Its dependence on
local errors and intervehicle information is specified in
Section~\ref{sec:platoon_control}.

With exact parameter knowledge, this is achieved by the
steering law
\begin{align}
\label{eq:exact_arc_fl}
    \delta_{f,i}(l_d)
    =&\frac{1}{g_{1,i}}
    \Big[
      -f_{1,i}(l_d)-d_{1,i}(l_d)\dot\theta_{R,i}(l_d)
      \notag \\ 
      &~~+\dot v_{x,i}(l_d)\left(e_{lat,i}^{ref}(l_d)\right)'
      +v_{x,i}^{2}(l_d)\mu_i(l_d)
    \Big].
\end{align}
Substituting \eqref{eq:exact_arc_fl} into
\eqref{eq:arc_length_io} gives
\eqref{eq:generic_arc_dynamics}. Thus, vehicle-specific
parameters and longitudinal-motion terms are compensated
through the steering law, allowing the subsequent platoon-level
design to be expressed through $\mu_i$ using the same nominal
lateral error dynamics for every vehicle.

During implementation, the controller projects the vehicle's
current position onto the desired path to determine the
corresponding arc-length coordinate $l_d$. It then evaluates
\eqref{eq:exact_arc_fl} at this coordinate to compute the
steering command applied at the current time.

Feedback linearization assigns the lateral error dynamics,
but these do not describe the vehicle's full state evolution.
The remaining state dynamics, referred to as the internal
dynamics, must also be stable. We retain their time-domain
representation. For each vehicle satisfying the
internal-stability conditions in \cite{ch3preprint},
evaluated using its own parameters and longitudinal motion,
the internal dynamics are input-to-state stable with respect
to $(e_{lat,i}^{ref},\dot e_{lat,i}^{ref},\dot\theta_{R,i})$.
Since
$\dot e_{lat,i}^{ref}\approx v_{x,i}(e_{lat,i}^{ref})'$
by \eqref{eq:arc_first_derivative}, bounded lateral
errors and their first spatial derivatives ensure bounded
internal states, provided the longitudinal speed and reference
heading rate are also bounded.

\subsection{Perturbed external dynamics}
\label{subsec:arc_perturbation}

The steering law \eqref{eq:exact_arc_fl} accounts for
heterogeneity through each vehicle's own parameter values,
but exact cancellation requires these values to be known
precisely. Parameter uncertainty introduces a mismatch
between the vehicle dynamics and the model used for
cancellation. Together with external disturbances and
discrepancies from the nominal model, this mismatch produces
an aggregate local perturbation, represented in the spatial
lateral error dynamics by
\begin{equation}
\label{eq:perturbed_arc_generic}
    \left(e_{lat,i}^{ref}(l_d)\right)''
    =\mu_i(l_d)+w_i(l_d),
\end{equation}
where $w_i(l_d)$ denotes the resulting local perturbation.

Vehicle-level robust control or disturbance compensation can be
used to mitigate this perturbation, for example through
incremental nonlinear dynamic inversion
\cite{ch3preprint}, disturbance-observer-based control
\cite{kayacan2019feedback,wang2022trajectory}, or active
disturbance rejection control
\cite{xia2016lateral,zhao2025lateral}.
When such compensation is used, $w_i$ in
\eqref{eq:perturbed_arc_generic} represents the residual
perturbation that remains after compensation. The subsequent
platoon-level analysis characterizes lateral error propagation
in terms of these local perturbations, without prescribing
a particular vehicle-level compensation method.

For brevity, the argument $l_d$ is omitted below when
unambiguous.

\section{Platoon Control Formulation under Different Information Modes}
\label{sec:platoon_control}
We now specify the reference path and the virtual input $\mu_i$
in \eqref{eq:perturbed_arc_generic} for two information modes.
With onboard sensing, each follower tracks its predecessor's
traveled path. With V2V communication, every vehicle tracks
the desired path, and predecessor information is additionally
used to construct a learned input.

\subsection{Onboard sensing mode: predecessor tracking}
\label{subsec:onboard_sensing}
In the onboard sensing mode, the lead vehicle uses the desired
path as its reference path. Each following vehicle uses onboard
sensing to record its immediate predecessor's traveled path
and subsequently tracks that path. We refer to this configuration
as predecessor tracking (PT).

\begin{figure}[h]
    \centering
    \subfloat[Predecessor tracking]{%
        \includegraphics[width=0.99\columnwidth]
        {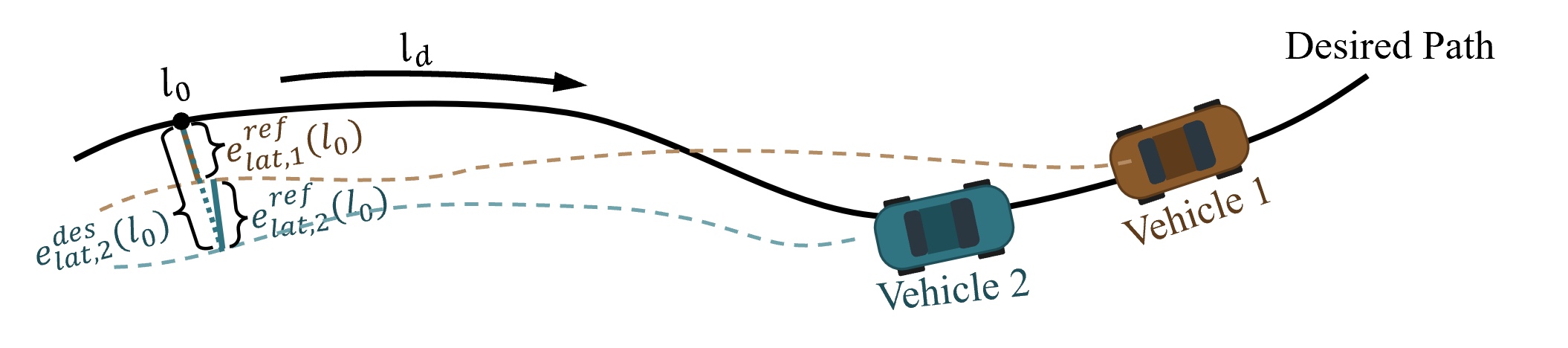}%
        \label{fig:pt_error_geometry}%
    }

    \medskip
    \subfloat[Desired-path tracking]{%
        \includegraphics[width=0.99\columnwidth]
        {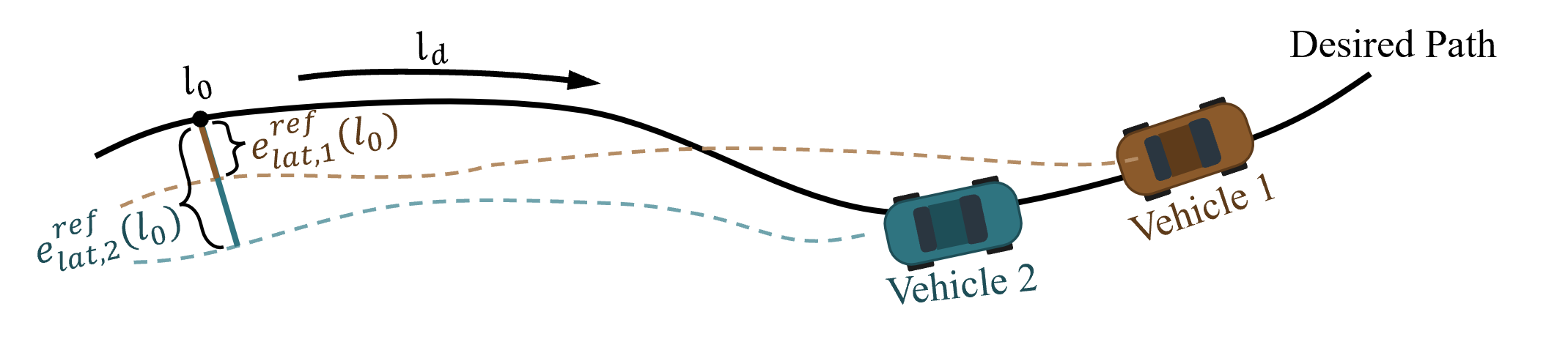}%
        \label{fig:dt_error_geometry}%
    }
    \caption{Reference paths and lateral errors under PT and DT.
    }
    \label{fig:tracking_configurations}
\end{figure}

Let $e_{lat,i}^{des}$ denote the lateral error of vehicle $i$
relative to the desired path. Whereas $e_{lat,i}^{ref}$ is used
for local tracking control, $e_{lat,i}^{des}$ describes the
vehicle's deviation from the path intended for the platoon.
Figure~\ref{fig:tracking_configurations}(a) illustrates this
distinction for two vehicles. Using the geometric approximation
in \cite{li2026lateral}, these errors are related by
\begin{equation}
\label{eq:pt_error_relation}
e_{lat,i}^{ref}
=
\begin{cases}
e_{lat,1}^{des}, & i=1,\\[1mm]
e_{lat,i}^{des}-e_{lat,i-1}^{des},
& i=2,\ldots,N.
\end{cases}
\end{equation}
This relation expresses that, at a common location $l_d$, a following
vehicle's deviation from its predecessor's traveled path is approximately
the difference between their respective deviations from the desired path.

Since each follower has access only to its predecessor's
traveled path, we choose the virtual input based on the
reference-path tracking error:
\begin{equation}
\label{eq:pt_mu}
\mu_i
=
-k_1e_{lat,i}^{ref}
-k_2\left(e_{lat,i}^{ref}\right)',
\qquad i=1,\ldots,N,
\end{equation}
where $k_1>0$ and $k_2>0$ are common controller gains. Substituting
\eqref{eq:pt_mu} into \eqref{eq:perturbed_arc_generic} gives
\begin{equation}
\label{eq:pt_ref_dynamics}
\left(e_{lat,i}^{ref}\right)''
+k_2\left(e_{lat,i}^{ref}\right)'
+k_1e_{lat,i}^{ref}
=
w_i.
\end{equation}
Using \eqref{eq:pt_error_relation}, the lead vehicle satisfies
\begin{equation}
\label{eq:pt_leader}
\left(e_{lat,1}^{des}\right)''
+k_2\left(e_{lat,1}^{des}\right)'
+k_1e_{lat,1}^{des}
=
w_1,
\end{equation}
whereas each following vehicle satisfies
\begin{align}
\label{eq:pt_follower}
&\left(e_{lat,i}^{des}-e_{lat,i-1}^{des}\right)''
+k_2\left(e_{lat,i}^{des}-e_{lat,i-1}^{des}\right)'
\notag\\
&\qquad
+k_1\left(e_{lat,i}^{des}-e_{lat,i-1}^{des}\right)
=w_i,
\qquad i=2,\ldots,N.
\end{align}

\subsection{V2V mode: learn-from-predecessor strategy}
\label{subsec:v2v}
In the V2V mode, the desired path is communicated throughout
the platoon, and every vehicle uses it as its reference path.
As illustrated in Fig.~\ref{fig:tracking_configurations}(b),
the reference lateral error therefore coincides with the
desired-path lateral error:
\begin{equation}
\label{eq:dt_error_relation}
e_{lat,i}^{ref}
=
e_{lat,i}^{des},
\qquad i=1,\ldots,N.
\end{equation}
Besides sharing the desired path, V2V communication allows
each follower to use information recorded during its
predecessor's traversal. We propose a learn-from-predecessor
(LFP) strategy for the feedback-linearized dynamics by
introducing a learned term into the virtual input:
\begin{equation}
\label{eq:lfp_mu}
    \mu_i
    =
    -k_1e_{lat,i}^{des}
    -k_2\left(e_{lat,i}^{des}\right)'
    +v_i,
    \qquad i=1,\ldots,N.
\end{equation}
The learned term $v_i(l_d)$ is generated recursively along
the platoon:
\begin{equation}
\label{eq:lfp_update}
\begin{aligned}
v_1 &\equiv 0,\\
v_i
&=v_{i-1}
+k_{LP}e_{lat,i-1}^{des}
+k_{LD}\left(e_{lat,i-1}^{des}\right)'\\
&\quad
+k_{LDD}\left(e_{lat,i-1}^{des}\right)'',
\qquad i=2,\ldots,N.
\end{aligned}
\end{equation}
Here, $k_{LP}$, $k_{LD}$, and $k_{LDD}$ are common learning
gains for the lateral error and its first and second spatial
derivatives, respectively. The lead vehicle has $v_1\equiv0$
because it has no predecessor. During its traversal, vehicle
$i-1$ records its lateral error and spatial derivatives,
combines them with its learned term $v_{i-1}$ according to
\eqref{eq:lfp_update}, and communicates the resulting $v_i$
to vehicle $i$. The follower uses this term when it reaches
the corresponding desired-path location.

Substituting \eqref{eq:lfp_mu} into \eqref{eq:perturbed_arc_generic} and
using \eqref{eq:dt_error_relation}, the lead vehicle, for which
$v_1\equiv0$, satisfies
\begin{equation}
\label{eq:v2v_leader}
    \left(e_{lat,1}^{des}\right)''
    +k_2\left(e_{lat,1}^{des}\right)'
    +k_1e_{lat,1}^{des}
    =
    w_1,
\end{equation}
and every following vehicle satisfies
\begin{equation}
\label{eq:v2v_follower}
    \left(e_{lat,i}^{des}\right)''
    +k_2\left(e_{lat,i}^{des}\right)'
    +k_1e_{lat,i}^{des}
    =
    v_i+w_i,
    \qquad i=2,\ldots,N.
\end{equation}
With $v_i\equiv0$ for every vehicle, the V2V formulation
reduces to independent desired-path tracking with proportional
and derivative feedback, referred to below as the
desired-path tracking (DT) baseline. The LFP strategy
supplements this baseline with the predecessor-dependent
update \eqref{eq:lfp_update}.

The intervehicle coupling therefore enters differently in
PT and LFP. In PT, each follower's reference path is its
predecessor's traveled path, leading to the error relation
\eqref{eq:pt_error_relation}. In LFP, every vehicle tracks
the desired path, while predecessor information enters
through the learned term in the virtual input.

\subsection{Sensing and communication delays}
\label{subsec:communication_delay}

\begin{figure}[h]
    \centering
    \includegraphics[width=0.9\columnwidth]
    {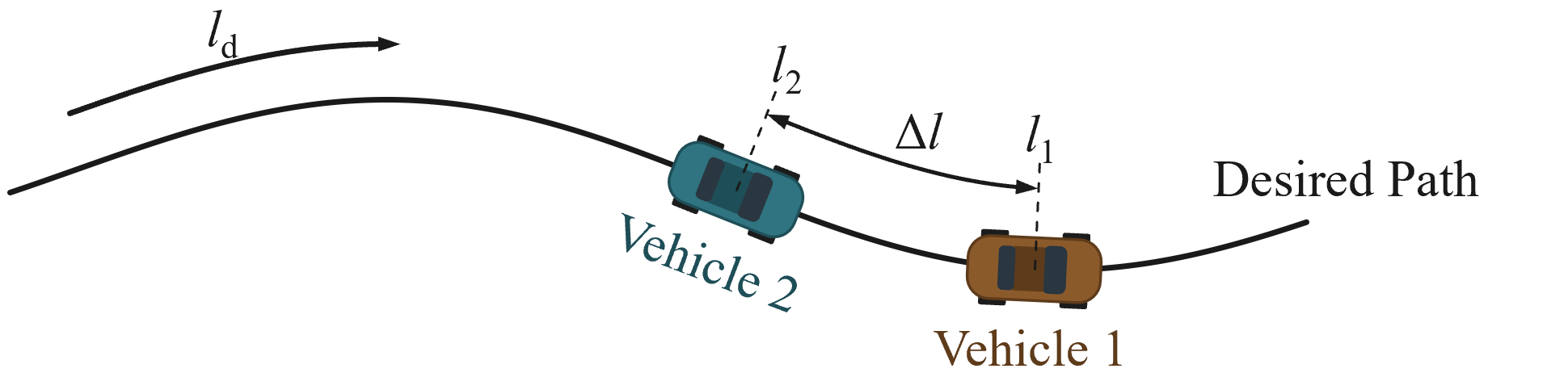}
    \caption{Spatial separation provides time for information
    acquisition. }
    \label{fig:spatial_time_gap}
\end{figure}

Spatial separation provides time to acquire predecessor
information before it is needed \cite{li2026lateral}.
In Fig.~\ref{fig:spatial_time_gap}, the predecessor is at
$l_1$ while the follower is at $l_2$, separated by
$\Delta l=l_1-l_2$. At approximately constant longitudinal
speed $v_f$, the follower reaches $l_1$ after approximately
$\Delta l/v_f$. Only then does it need the predecessor's
path at $l_1$ in PT or the learned term associated with
$l_1$ in LFP. Thus, the control law is unaffected provided
the combined sensing, processing, and communication delay
is shorter than $\Delta l/v_f$.

\section{Robust Lateral String Stability}
\label{sec:robust_lateral_string_stability}
The local perturbations $w_i$ in
\eqref{eq:perturbed_arc_generic} affect individual vehicles
and propagate through their interconnections. We therefore
introduce a robust $\mathcal L_2$ lateral string stability
notion that accounts for both initial conditions and
perturbations throughout the platoon. We then establish
a sufficient condition for a class of one-predecessor LTI
interconnections and derive a corresponding
$\mathcal L_\infty$ bound on lateral errors.

Consider a platoon of $N$ interconnected systems evolving
with respect to the desired-path arc length
$l_d\in[0,\infty)$. Let $x_{e,i}(l_d)$ denote the error state of vehicle $i$,
$y_i(l_d)$ a specified lateral performance output, and
$w_i(l_d)$ a local perturbation. The following
definition imposes no particular form on the subsystem
dynamics or interconnection structure. All $\mathcal L_2$
and $\mathcal L_\infty$ norms below are taken with respect
to $l_d$ over $[0,\infty)$, with
$\|q\|_{\mathcal L_2}
=(\int_0^\infty\|q(l_d)\|_2^2\,dl_d)^{1/2}$ and
$\|q\|_{\mathcal L_\infty}
=\sup_{l_d\ge0}\|q(l_d)\|_2$.
Here, $\|\cdot\|_2$ denotes the Euclidean norm on vectors
and the induced $2$-norm on matrices.

\subsection{Definition}
\label{subsec:robust_lss_definition}
Recall that a function
$\varphi:\mathbb R_{\ge0}\rightarrow\mathbb R_{\ge0}$ is of class
$\mathcal K_\infty$ if it is continuous, strictly increasing,
$\varphi(0)=0$, and $\varphi(r)\rightarrow\infty$ as
$r\rightarrow\infty$.
\begin{definition}[Robust $\mathcal L_2$ lateral string stability]
\label{def:robust_l2_lss}
For a specified lateral performance output $y_i$, the platoon is
robustly $\mathcal L_2$ lateral string stable if there exist functions
$\rho_2,\sigma_2\in\mathcal K_\infty$, independent of the platoon
length $N$, such that
\begin{equation}
\label{eq:robust_l2_lss}
    \max_{1\le i\le N}
    \|y_i\|_{\mathcal L_2}
    \le
    \rho_2\Big(
                \max_{1\le i\le N}\|x_{e,i}(0)\|_2
    \Big)
    +
    \sigma_2\Big(
        \max_{1\le i\le N}\|w_i\|_{\mathcal L_2}
    \Big)
\end{equation}
for every platoon length $N$, every set of initial
conditions, and every collection of local perturbations
$w_i\in\mathcal L_2$.
\end{definition}
The independence of $\rho_2$ and $\sigma_2$ from $N$
expresses scalability with respect to platoon length:
if the initial-state and local perturbation norms remain
uniformly bounded as vehicles are added, the
$\mathcal L_2$ output norms throughout the platoon must
remain bounded by the same bound.
\begin{rem}
\label{rem:definition_relation}
The $\mathcal L_2$ lateral string stability notion in
\cite{li2026lateral} requires contraction of the
predecessor-to-follower output map under zero initial
conditions and without local perturbations.
Definition~\ref{def:robust_l2_lss} instead requires
platoon-length-uniform bounds in the presence of initial
errors and distributed perturbations. Downstream error
attenuation is examined separately in
Section~\ref{sec:configuration_analysis}.
\end{rem}
Although the definition is stated over $l_d\in[0,\infty)$, a practical
driving operation takes place over a finite path length. Consequently, any bounded perturbation acting over such an operation,
when extended by zero outside the operating interval, has a finite
spatial $\mathcal L_2$ norm. The $\mathcal L_2$ formulation therefore
provides a natural measure of the accumulated effect of errors over a driving operation.

For the lateral-control problem considered here, the
performance output is chosen as the desired-path lateral
error, $y_i=e_{lat,i}^{des}$, the error state is
$x_{e,i}=[\,e_{lat,i}^{des},\ (e_{lat,i}^{des})'\,]^T$,
and $w_i$ is the local perturbation introduced in
\eqref{eq:perturbed_arc_generic}.

\subsection{General conditions for LTI interconnections}
\label{subsec:general_lti_conditions}
We next derive a sufficient condition under which a general one-predecessor
LTI interconnection admits a platoon-length-uniform $\mathcal L_2$ bound of
the form appearing in Definition~\ref{def:robust_l2_lss}. Let $y_i$ denote
the performance output of subsystem $i$. Consider
\begin{equation}
\label{eq:general_lti_leader}
x_1'=Ax_1+B_\psi\psi_1,
\qquad y_1=C_yx_1.
\end{equation}
\begin{equation}
\label{eq:general_lti_follower}
\begin{aligned}
x_i'&=Ax_i+B_py_{i-1}+B_\psi\psi_i,\\
y_i&=C_yx_i+D_py_{i-1},
\qquad i=2,\ldots,N.
\end{aligned}
\end{equation}
Here, $x_i\in\mathbb R^n$ and
$\psi_i\in\mathbb R^{n_\psi}$ are the state and local
input of the state-space realization for vehicle $i$,
and $y_i\in\mathbb R^{n_y}$ is its performance output.
The state and signal dimensions are arbitrary but fixed,
and $A$, $B_p$, $B_\psi$, $C_y$, and $D_p$ are independent
of both the vehicle index and the platoon length.
The feedthrough term $D_py_{i-1}$ allows the
predecessor-to-follower map to be proper rather than
strictly proper. The realization state $x_i$ and local
input $\psi_i$ need not coincide with the error state
$x_{e,i}$ and physical perturbation $w_i$ used in
Definition~\ref{def:robust_l2_lss}. Their relationship
is addressed in Remark~\ref{rem:psi_w}.

For zero initial conditions, taking the Laplace transform with respect to
$l_d$ gives, for each follower,
\begin{equation}
\label{eq:general_lti_transfer}
    Y_i(s)
    =
    H(s)Y_{i-1}(s)
    +
    T_{y\psi}(s)\Psi_i(s),
    \qquad i=2,\ldots,N,
\end{equation}
where
\begin{equation}
\label{eq:general_transfer_functions}
    H(s)
    =
    D_p+C_y(sI-A)^{-1}B_p,
    \quad
    T_{y\psi}(s)
    =
    C_y(sI-A)^{-1}B_\psi.
\end{equation}
Here, $H$ is the predecessor-to-follower transfer matrix of the
performance output, while $T_{y\psi}$ is the local-input-to-output
transfer matrix. For the leader,
\begin{equation}
\label{eq:general_lti_leader_transfer}
    Y_1(s)=T_{y\psi}(s)\Psi_1(s).
\end{equation}

For a stable transfer matrix $G$, its $\mathcal H_\infty$
norm is
$\|G\|_{\mathcal H_\infty}
=\sup_{\omega\in\mathbb R}\bar\sigma(G(j\omega))$,
where $\bar\sigma$ denotes the largest singular value.
This norm equals the induced $\mathcal L_2$ gain.
\begin{thm}[Sufficient condition for robust $\mathcal L_2$ lateral string stability]
\label{thm:general_l2_condition}
Suppose that $A$ is Hurwitz and
\begin{equation}
\label{eq:general_hinf_contraction}
    \|H\|_{\mathcal H_\infty}<1.
\end{equation}
Then there exist $\rho_2,\sigma_2\in\mathcal K_\infty$, independent of the
platoon length $N$, such that the LTI interconnection
\eqref{eq:general_lti_leader}--\eqref{eq:general_lti_follower} satisfies
\begin{equation}
\label{eq:general_l2_bound}
    \max_{1\le i\le N}\|y_i\|_{\mathcal L_2}
    \le
    \rho_2\Big(\max_{1\le i\le N}\|x_i(0)\|_2\Big)
    +
    \sigma_2\Big(\max_{1\le i\le N}\|\psi_i\|_{\mathcal L_2}\Big).
\end{equation}
\end{thm}

\begin{proof}
Since $A$ is Hurwitz, there exist constants $M>0$ and $\lambda>0$
such that
\begin{equation}
\label{eq:general_exponential_bound}
    \|e^{Al_d}\|_2
    \le
    Me^{-\lambda l_d},
    \qquad l_d\ge0.
\end{equation}
Using \eqref{eq:general_exponential_bound}, the contribution of the
local initial condition to the performance output satisfies
\begin{align}
    \left\|
        C_ye^{Al_d}x_i(0)
    \right\|_{\mathcal L_2}^2
    &=
    \int_0^\infty
    \left\|
        C_ye^{A\ell}x_i(0)
    \right\|_2^2\,d\ell
    \notag\\
    &\le
    \int_0^\infty
    \|C_y\|_2^2
    \|e^{A\ell}\|_2^2
    \|x_i(0)\|_2^2\,d\ell
    \notag\\
    &\le
    \|C_y\|_2^2M^2
    \|x_i(0)\|_2^2
    \int_0^\infty
    e^{-2\lambda\ell}\,d\ell
    \notag\\
    &=
    \tfrac{\|C_y\|_2^2M^2}{2\lambda}
    \|x_i(0)\|_2^2.
\end{align}
Taking square roots gives
\begin{equation}
\label{eq:general_local_initial_bound}
    \left\|
        C_ye^{Al_d}x_i(0)
    \right\|_{\mathcal L_2}
    \le
    \tfrac{\|C_y\|_2M}{\sqrt{2\lambda}}
    \|x_i(0)\|_2.
\end{equation}
The solution of \eqref{eq:general_lti_leader} gives
$y_1(l_d)=C_ye^{Al_d}x_1(0)
+\int_0^{l_d}C_ye^{A(l_d-\ell)}B_\psi\psi_1(\ell)\,d\ell$.
Using \eqref{eq:general_local_initial_bound} together with the induced
$\mathcal L_2$ gain of $T_{y\psi}$,
\begin{equation}
\label{eq:general_y_leader_bound}
    \|y_1\|_{\mathcal L_2}
    \le
    \tfrac{\|C_y\|_2M}{\sqrt{2\lambda}}
    \|x_1(0)\|_2
    +
    \|T_{y\psi}\|_{\mathcal H_\infty}
    \|\psi_1\|_{\mathcal L_2}.
\end{equation}
For $i=2,\ldots,N$, the solution of \eqref{eq:general_lti_follower},
together with its output equation $y_i=C_yx_i+D_py_{i-1}$, gives
\begin{align}
\label{eq:general_y_solution}
    y_i(l_d)
    ={}&
    C_ye^{Al_d}x_i(0)
    +
    D_py_{i-1}(l_d) \notag \\
    &+
    \int_0^{l_d}
    C_ye^{A(l_d-\ell)}B_py_{i-1}(\ell)\,d\ell\notag \\
    &
    +
    \int_0^{l_d}
    C_ye^{A(l_d-\ell)}B_\psi\psi_i(\ell)\,d\ell.
\end{align}
The feedthrough term and the predecessor convolution together constitute the
proper operator $H$, whose induced $\mathcal L_2$ gain is
$\|H\|_{\mathcal H_\infty}$. Using \eqref{eq:general_local_initial_bound} and the induced
$\mathcal L_2$ gains of $H$ and $T_{y\psi}$ in
\eqref{eq:general_y_solution} gives
\begin{align}
\label{eq:general_y_recursion}
    \|y_i\|_{\mathcal L_2}
    \le &
    \tfrac{\|C_y\|_2M}{\sqrt{2\lambda}}
    \|x_i(0)\|_2
    +
    \|H\|_{\mathcal H_\infty}
    \|y_{i-1}\|_{\mathcal L_2} \notag \\
    &+
    \|T_{y\psi}\|_{\mathcal H_\infty}
    \|\psi_i\|_{\mathcal L_2}.
\end{align}
Iterating \eqref{eq:general_y_recursion} from
\eqref{eq:general_y_leader_bound} yields, for each
$i=1,\ldots,N$,
\begin{align}
\label{eq:general_y_recursive_bound}
    \|y_i\|_{\mathcal L_2}
    \le{}&
    \tfrac{\|C_y\|_2M}{\sqrt{2\lambda}}
    \sum_{j=1}^{i}
    \|H\|_{\mathcal H_\infty}^{\,i-j}
    \|x_j(0)\|_2 \notag \\
    &+
    \|T_{y\psi}\|_{\mathcal H_\infty}
    \sum_{j=1}^{i}
    \|H\|_{\mathcal H_\infty}^{\,i-j}
    \|\psi_j\|_{\mathcal L_2}.
\end{align}
Using
\begin{equation}
\label{eq:general_max_bounds}
    \|x_j(0)\|_2
    \le
    \max_{1\le k\le N}\|x_k(0)\|_2,
    \qquad
    \|\psi_j\|_{\mathcal L_2}
    \le
    \max_{1\le k\le N}\|\psi_k\|_{\mathcal L_2},
\end{equation}
in \eqref{eq:general_y_recursive_bound}, and using
\begin{equation}
\label{eq:general_geometric_sum}
    \sum_{j=1}^{i}
    \|H\|_{\mathcal H_\infty}^{\,i-j}
    =
    \tfrac{
        1-\|H\|_{\mathcal H_\infty}^{\,i}
    }{
        1-\|H\|_{\mathcal H_\infty}
    },
\end{equation}
gives
\begin{align}
    \|y_i\|_{\mathcal L_2}
    \le{}&
    \tfrac{\|C_y\|_2M}{\sqrt{2\lambda}}
    \tfrac{
        1-\|H\|_{\mathcal H_\infty}^{\,i}
    }{
        1-\|H\|_{\mathcal H_\infty}
    }
    \max_{1\le k\le N}\|x_k(0)\|_2 \notag\\
    &+
    \|T_{y\psi}\|_{\mathcal H_\infty}
    \tfrac{
        1-\|H\|_{\mathcal H_\infty}^{\,i}
    }{
        1-\|H\|_{\mathcal H_\infty}
    }
    \max_{1\le k\le N}\|\psi_k\|_{\mathcal L_2}.
\end{align}
Since $\|H\|_{\mathcal H_\infty}<1$,
\begin{align}
\label{eq:general_l2_explicit_bound}
    \max_{1\le i\le N}
    \|y_i\|_{\mathcal L_2}
    \le{}&
    \tfrac{
        \|C_y\|_2M
    }{
        \sqrt{2\lambda}
        \left(1-\|H\|_{\mathcal H_\infty}\right)
    }
    \max_{1\le i\le N}\|x_i(0)\|_2 \notag \\
    &+
    \tfrac{
        \|T_{y\psi}\|_{\mathcal H_\infty}
    }{
        1-\|H\|_{\mathcal H_\infty}
    }
    \max_{1\le i\le N}\|\psi_i\|_{\mathcal L_2}.
\end{align}
Let $a_x$ and $a_\psi$ denote, respectively, the two coefficients
multiplying the initial-condition and local-input terms in
\eqref{eq:general_l2_explicit_bound}. Since
$\|H\|_{\mathcal H_\infty}<1$, both coefficients are finite, nonnegative,
and independent of $N$. Choosing any constants
$\bar a_x>a_x$ and $\bar a_\psi>a_\psi$ gives \eqref{eq:general_l2_bound}
with $\rho_2(r)=\bar a_x r$ and $\sigma_2(r)=\bar a_\psi r$, both in
$\mathcal K_\infty$.
\end{proof}
\begin{rem}
\label{rem:psi_w}
Theorem~\ref{thm:general_l2_condition} bounds the performance
output in terms of the state $x_i$ and local input $\psi_i$
of the state-space realization
\eqref{eq:general_lti_leader}--\eqref{eq:general_lti_follower},
whereas Definition~\ref{def:robust_l2_lss} is stated in
terms of the error state $x_{e,i}$ and physical
perturbation $w_i$. Applying the theorem to establish
the definition therefore requires
\begin{align}
\label{eq:psi_w_bridge}
\max_{1\le i\le N}\|\psi_i\|_{\mathcal L_2}
&\le
\gamma_\psi\Big(
    \max_{1\le i\le N}\|w_i\|_{\mathcal L_2}
\Big),
\notag\\
\max_{1\le i\le N}\|x_i(0)\|_2
&\le
\gamma_x\Big(
    \max_{1\le i\le N}\|x_{e,i}(0)\|_2
\Big),
\end{align}
for some $\gamma_\psi,\gamma_x\in\mathcal K_\infty$ independent of $N$. The
compositions of these with the functions $\rho_2$ and $\sigma_2$ of
\eqref{eq:general_l2_bound} are again of class $\mathcal K_\infty$, so
$\rho_2\circ\gamma_x$ and $\sigma_2\circ\gamma_\psi$ serve as the functions
required by \eqref{eq:robust_l2_lss}. For the LFP formulation, the perturbation relation in
Corollary~\ref{cor:v2v_robust_l2} gives
$\gamma_\psi(r)=2r$, while
Appendix~\ref{app:v2v_realization} establishes a
$\gamma_x$ depending only on the
controller gains.
\end{rem}

Theorem~\ref{thm:general_l2_condition} bounds the accumulated
output magnitude in the $\mathcal L_2$ sense. For lateral
path following, the peak lateral error is also important
because it measures the largest deviation from the desired
path and is therefore relevant to maintaining clearance
from obstacles and road boundaries.
For the LTI interconnection considered here, the same
conditions further imply a platoon-length-uniform
$\mathcal L_\infty$ output bound. With
$y_i=e_{lat,i}^{des}$, this bounds the peak desired-path
lateral error of every vehicle.
\begin{cor}[Uniform $\mathcal L_\infty$ output boundedness]
\label{cor:uniform_linf_bound}
Under the conditions of Theorem~\ref{thm:general_l2_condition}, there
exist constants $c_{\infty,x}>0$ and $c_{\infty,\psi}>0$, independent of
the platoon length $N$, such that
\begin{equation}
\label{eq:general_linf_bound}
    \max_{1\le i\le N}
    \|y_i\|_{\mathcal L_\infty}
    \le
    c_{\infty,x}
    \max_{1\le i\le N}\|x_i(0)\|_2
    +
    c_{\infty,\psi}
    \max_{1\le i\le N}\|\psi_i\|_{\mathcal L_2}.
\end{equation}
\end{cor}
\begin{proof}
Let
\begin{equation}
\label{eq:general_y_impulse_responses}
    h_{yy}(l_d)=C_ye^{Al_d}B_p,
    \qquad
    h_{y\psi}(l_d)=C_ye^{Al_d}B_\psi.
\end{equation}
Since $A$ is Hurwitz, both $h_{yy}$ and $h_{y\psi}$ decay exponentially
and therefore belong to $\mathcal L_2$. 
Moreover, \eqref{eq:general_exponential_bound} gives
\begin{equation}
\label{eq:general_CyeA_bound}
    \sup_{l_d\ge0}
    \|C_ye^{Al_d}\|_2
    \le
    \|C_y\|_2M.
\end{equation}
The term $C_y(sI-A)^{-1}B_p$ is strictly proper, so $H(j\omega)\to D_p$ as
$\omega\to\infty$ and therefore
\begin{equation}
\label{eq:general_Dp_bound}
    \|D_p\|_2
    =
    \lim_{\omega\to\infty}\|H(j\omega)\|_2
    \le
    \|H\|_{\mathcal H_\infty}
    <1.
\end{equation}
For each follower, the solution of
\eqref{eq:general_lti_follower} gives
\begin{align}
\label{eq:general_follower_y_solution}
    y_i(l_d)
    ={}&
    C_ye^{Al_d}x_i(0)
    +
    D_py_{i-1}(l_d)
    \notag\\
    &+
    \int_0^{l_d}
    h_{yy}(l_d-\ell)y_{i-1}(\ell)\,d\ell \notag \\
    &+
    \int_0^{l_d}
    h_{y\psi}(l_d-\ell)\psi_i(\ell)\,d\ell,
    \qquad i=2,\ldots,N.
\end{align}
For the first convolution term, the triangle inequality for integrals
and compatibility of the induced matrix 2-norm with the Euclidean vector norm, followed by the
Cauchy--Schwarz inequality, give, \begin{align}
\label{eq:general_hyy_convolution_bound}
    &~~~~\Big\|
        \int_0^{l_d}
        h_{yy}(l_d-\ell)y_{i-1}(\ell)\,d\ell
    \Big\|_2 \notag\\
    &\le
    \int_0^{l_d}
    \|h_{yy}(l_d-\ell)\|_2
    \|y_{i-1}(\ell)\|_2\,d\ell
    \notag\\
    &\le
    \Big(
        \int_0^{l_d}
        \|h_{yy}(l_d-\ell)\|_2^2\,d\ell
    \Big)^{1/2}
    \Big(
        \int_0^{l_d}
        \|y_{i-1}(\ell)\|_2^2\,d\ell
    \Big)^{1/2}
    \notag\\
    &\le
    \|h_{yy}\|_{\mathcal L_2}
    \|y_{i-1}\|_{\mathcal L_2},
    \qquad \forall\,l_d\ge0.
\end{align}
The last inequality follows because, after the change of
variables $u=l_d-\ell$ in the first integral, both integrals
over $[0,l_d]$ are bounded by the corresponding integrals
over $[0,\infty)$. Similarly,
\begin{equation}
\label{eq:general_hypsi_convolution_bound}
\begin{aligned}
    &\Big\|
        \int_0^{l_d}
        h_{y\psi}(l_d-\ell)\psi_i(\ell)\,d\ell
    \Big\|_2\le
    \|h_{y\psi}\|_{\mathcal L_2}
    \|\psi_i\|_{\mathcal L_2},
    \quad \forall\,l_d\ge0.
\end{aligned}
\end{equation}
Applying the triangle inequality to
\eqref{eq:general_follower_y_solution} and using
\eqref{eq:general_CyeA_bound},
\eqref{eq:general_hyy_convolution_bound}, and
\eqref{eq:general_hypsi_convolution_bound} gives
\begin{align}
    \|y_i(l_d)\|_2
    \le{}&
    \|D_p\|_2\|y_{i-1}(l_d)\|_2
    +\|C_y\|_2M\|x_i(0)\|_2 \notag\\
    &+\|h_{yy}\|_{\mathcal L_2}\|y_{i-1}\|_{\mathcal L_2} \notag\\
    &+\|h_{y\psi}\|_{\mathcal L_2}\|\psi_i\|_{\mathcal L_2},
    \qquad \forall\,l_d\ge0.
\end{align}
Only the first term on the right depends on
$l_d$. Taking the supremum over $l_d\ge0$ therefore gives,
for $i=2,\ldots,N$,
\begin{align}
\label{eq:general_follower_linf_bound}
    \|y_i\|_{\mathcal L_\infty}
    \le&
    \|D_p\|_2
    \|y_{i-1}\|_{\mathcal L_\infty}
    +
    \|C_y\|_2M
    \|x_i(0)\|_2 \notag \\
    &+
    \|h_{yy}\|_{\mathcal L_2}
    \|y_{i-1}\|_{\mathcal L_2}
    +
    \|h_{y\psi}\|_{\mathcal L_2}
    \|\psi_i\|_{\mathcal L_2}.
\end{align}
For the leader,
\begin{equation}
\label{eq:general_leader_y_solution}
    y_1(l_d)
    =
    C_ye^{Al_d}x_1(0)
    +
    \int_0^{l_d}
    h_{y\psi}(l_d-\ell)\psi_1(\ell)\,d\ell.
\end{equation}
Using \eqref{eq:general_CyeA_bound} and
\eqref{eq:general_hypsi_convolution_bound} in
\eqref{eq:general_leader_y_solution}, and taking the supremum over
$l_d\ge0$, gives
\begin{equation}
\label{eq:general_leader_linf_bound}
    \|y_1\|_{\mathcal L_\infty}
    \le
    \|C_y\|_2M
    \|x_1(0)\|_2
    +
    \|h_{y\psi}\|_{\mathcal L_2}
    \|\psi_1\|_{\mathcal L_2}.
\end{equation}
For every follower,
\[
    \|y_{i-1}\|_{\mathcal L_2}
    \le
    \max_{1\le j\le N}
    \|y_j\|_{\mathcal L_2}.
\]
Substituting the uniform $\mathcal L_2$ bound
\eqref{eq:general_l2_explicit_bound} into
\eqref{eq:general_follower_linf_bound}, and bounding the local
initial-condition and local-input terms by their corresponding
platoon-wide maxima, gives
\begin{equation}
\label{eq:general_linf_recursion}
    \|y_i\|_{\mathcal L_\infty}
    \le
    \|D_p\|_2
    \|y_{i-1}\|_{\mathcal L_\infty}
    +
    b,
    \qquad i=2,\ldots,N,
\end{equation}
where
\begin{align}
\label{eq:general_linf_offset}
    b
    :={}&
    \left[
        \|C_y\|_2M
        +
        \tfrac{
            \|h_{yy}\|_{\mathcal L_2}
            \|C_y\|_2M
        }{
            \sqrt{2\lambda}
            \left(1-\|H\|_{\mathcal H_\infty}\right)
        }
    \right]
    \max_{1\le j\le N}
    \|x_j(0)\|_2
    \notag\\
    &+
    \left[
        \tfrac{
            \|h_{yy}\|_{\mathcal L_2}
            \|T_{y\psi}\|_{\mathcal H_\infty}
        }{
            1-\|H\|_{\mathcal H_\infty}
        }
        +
        \|h_{y\psi}\|_{\mathcal L_2}
    \right]
    \max_{1\le j\le N}
    \|\psi_j\|_{\mathcal L_2}.
\end{align}
The quantity $b$ is independent of the vehicle index, and the leader
bound \eqref{eq:general_leader_linf_bound} gives
$\|y_1\|_{\mathcal L_\infty}\le b$. Iterating
\eqref{eq:general_linf_recursion} from this base case yields
\[
    \|y_i\|_{\mathcal L_\infty}
    \le
    b
    \sum_{j=0}^{i-1}
    \|D_p\|_2^{\,j},
    \qquad i=1,\ldots,N,
\]
and, since $\|D_p\|_2<1$ by \eqref{eq:general_Dp_bound}, the geometric
series is bounded uniformly in $i$, giving
\begin{align}
\label{eq:general_linf_explicit_bound}
    &\max_{1\le i\le N}
    \|y_i\|_{\mathcal L_\infty} \notag \\
    \le{}&
    \tfrac{1}{1-\|D_p\|_2}
    \left[
        \|C_y\|_2M
        +
        \tfrac{
            \|h_{yy}\|_{\mathcal L_2}
            \|C_y\|_2M
        }{
            \sqrt{2\lambda}
            \left(1-\|H\|_{\mathcal H_\infty}\right)
        }
    \right]
    \max_{1\le i\le N}
    \|x_i(0)\|_2
    \notag\\
    &+
    \tfrac{1}{1-\|D_p\|_2}
    \left[
        \tfrac{
            \|h_{yy}\|_{\mathcal L_2}
            \|T_{y\psi}\|_{\mathcal H_\infty}
        }{
            1-\|H\|_{\mathcal H_\infty}
        }
        +
        \|h_{y\psi}\|_{\mathcal L_2}
    \right]
    \max_{1\le i\le N}
    \|\psi_i\|_{\mathcal L_2}.
\end{align}
The two coefficients are finite, nonnegative, and independent of $N$.
Choosing $c_{\infty,x}>0$ and $c_{\infty,\psi}>0$ no smaller than the
corresponding coefficients establishes \eqref{eq:general_linf_bound}.
\end{proof}

\section{Lateral String Stability Analysis under the Two Information Modes}
\label{sec:configuration_analysis}
We now analyze robust lateral string stability for the control
formulations in Section~\ref{sec:platoon_control}. We first
establish the limitation of PT under onboard sensing, then
derive learning-gain conditions for LFP using V2V to satisfy
Theorem~\ref{thm:general_l2_condition}. An explicit gain choice
further allows us to characterize downstream error attenuation
for LFP.
\subsection{Onboard sensing: predecessor tracking}
\label{subsec:onboard_analysis}
We first examine the onboard sensing mode under the robust
$\mathcal L_2$ lateral string stability definition.
\begin{prop}[Limitation of the PT formulation]
\label{prop:onboard_not_robust_l2}
The platoon governed by
\eqref{eq:pt_leader}--\eqref{eq:pt_follower}, with
performance output $y_i=e_{lat,i}^{des}$, cannot be
robustly $\mathcal L_2$ lateral string stable for any
choice of stabilizing controller gains $k_1,k_2>0$.
\end{prop}
\begin{proof}
Take zero initial conditions and let $w_i=d$ for every
vehicle, where $d\in\mathcal L_2$ is nonzero. Let $z$
be the zero-initial-condition solution of
\begin{equation}
\label{eq:pt_common_local_response}
    z''+k_2z'+k_1z=d.
\end{equation}
Since $k_1,k_2>0$, $z\in\mathcal L_2$ and
$\|z\|_{\mathcal L_2}>0$.
Equations~\eqref{eq:pt_leader}--\eqref{eq:pt_follower}
then give $e_{lat,1}^{des}=z$ and
$e_{lat,i}^{des}-e_{lat,i-1}^{des}=z$, hence
\begin{equation}
\label{eq:pt_linear_error_accumulation}
    e_{lat,i}^{des}=iz,
    \qquad i=1,\ldots,N.
\end{equation}
Definition~\ref{def:robust_l2_lss} would therefore require
$N\|z\|_{\mathcal L_2}\le
\sigma_2(\|d\|_{\mathcal L_2})$ for every $N$.
This is impossible because the right-hand side is
independent of $N$.
\end{proof}
Under the zero initial conditions and common perturbation
chosen in the proof, the reference lateral errors are
identical. They accumulate through
\eqref{eq:pt_error_relation}, yielding
\eqref{eq:pt_linear_error_accumulation} despite stable
individual tracking dynamics.

\subsection{V2V communication: learn-from-predecessor}
\label{subsec:v2v_analysis}
We next consider the V2V mode, first establishing conditions on the learning
gains under which the LFP interconnection satisfies
Theorem~\ref{thm:general_l2_condition}, and then examining a gain choice that
makes the predecessor-to-follower transfer function constant.

\textit{\textbf{Conditions on the learning gains:}}
Throughout the following analysis, $k_1>0$ and $k_2>0$. Laplace transforms
$E_i^{des}$ and $W_i$ of $e_{lat,i}^{des}$ and $w_i$ are taken with respect
to $l_d$, and transfer relations are stated for zero initial conditions.
Subtracting the predecessor's error dynamics from those of
vehicle $i$ in
\eqref{eq:v2v_leader}--\eqref{eq:v2v_follower} and substituting
$v_i-v_{i-1}$ from \eqref{eq:lfp_update} gives
\begin{equation}
\label{eq:lfp_recursion}
\begin{aligned}
&\quad~\left(e_{lat,i}^{des}\right)''
+k_2\left(e_{lat,i}^{des}\right)'
+k_1e_{lat,i}^{des}\\
&=
\left(1+k_{LDD}\right)\left(e_{lat,i-1}^{des}\right)''
+\left(k_2+k_{LD}\right)\left(e_{lat,i-1}^{des}\right)'\\
&\quad~
+\left(k_1+k_{LP}\right)e_{lat,i-1}^{des}
+w_i-w_{i-1},
\qquad i=2,\ldots,N.
\end{aligned}
\end{equation}
This recursion relates the desired-path lateral errors of
two consecutive vehicles, with the local perturbations
entering through $w_i-w_{i-1}$. Its realization in the
form of \eqref{eq:general_lti_leader}--%
\eqref{eq:general_lti_follower} is given in
Appendix~\ref{app:v2v_realization}.

Let $D(s):=s^2+k_2s+k_1$ be the closed-loop polynomial of
\eqref{eq:v2v_leader}--\eqref{eq:v2v_follower} and collect the learning
gains of \eqref{eq:lfp_update} in
\begin{equation}
\label{eq:lfp_KL}
    K_L(s)
    :=
    k_{LP}
    +sk_{LD}
    +s^2k_{LDD}.
\end{equation}
The right-hand coefficients of \eqref{eq:lfp_recursion} are then those of
$D(s)+K_L(s)$, so taking the Laplace transform of \eqref{eq:lfp_recursion}
and solving for $E_i^{des}(s)$ gives, for $i=2,\ldots,N$,
\begin{equation}
\label{eq:v2v_transfer_recursion}
    E_i^{des}(s)
    =
    H(s)E_{i-1}^{des}(s)
    +
    T_{y\psi}(s)\left(W_i(s)-W_{i-1}(s)\right),
\end{equation}
where $T_{y\psi}(s)=1/D(s)$ and the predecessor-to-follower transfer
function of the performance output $e_{lat,i}^{des}$ is
\begin{equation}
\label{eq:v2v_H}
    H(s)
    =
    \tfrac{D(s)+K_L(s)}{D(s)}.
\end{equation}

It remains to determine when the predecessor-to-follower transfer function
satisfies $\|H\|_{\mathcal H_\infty}<1$. Since $k_1>0$ and $k_2>0$, $D$ is
Hurwitz and $|D(j\omega)|>0$ for every $\omega\in\mathbb R$, so from
\eqref{eq:v2v_H}, for $\omega\in\mathbb R$,
\begin{equation}
\label{eq:v2v_frequency_inequality}
    \left|H(j\omega)\right|<1
    \iff
    \left|D(j\omega)\right|^2
    -\left|D(j\omega)+K_L(j\omega)\right|^2
    >0.
\end{equation}
Both terms are polynomials in $\omega^2$ of degree at most two, so their
difference is
\begin{equation}
\label{eq:v2v_certificate_polynomial}
    \left|D(j\omega)\right|^2
    -\left|D(j\omega)+K_L(j\omega)\right|^2
    =
    b_4\omega^4+b_2\omega^2+b_0,
\end{equation}
where, by expansion,
\begin{align}
\label{eq:v2v_certificate_coefficients}
    b_4&=-k_{LDD}\left(k_{LDD}+2\right),
    \notag\\
    b_2&=2k_{LP}
    +2k_{LDD}\left(k_{LP}+k_1\right)
    -k_{LD}^2
    -2k_2k_{LD},
    \notag\\
    b_0&=-k_{LP}\left(k_{LP}+2k_1\right).
\end{align}
Two necessary conditions follow directly. Because
$\|H\|_{\mathcal H_\infty}$ is a supremum over $\omega$, it is at least
$|H(0)|$ and at least $\lim_{\omega\rightarrow\infty}|H(j\omega)|$, and
\eqref{eq:v2v_H} gives $H(0)=1+k_{LP}/k_1$ and $H(\infty)=1+k_{LDD}$.
Hence $\|H\|_{\mathcal H_\infty}<1$ requires $-2k_1<k_{LP}<0$ and
$-2<k_{LDD}<0$.

An exact condition is obtained from
\eqref{eq:v2v_certificate_polynomial} by setting
$z=\omega^2\ge0$. Under the necessary conditions
$b_4>0$ and $b_0>0$, the quadratic
$b_4z^2+b_2z+b_0$ has minimum $b_0$ on $z\ge0$
when $b_2\ge0$, and minimum
$b_0-b_2^2/(4b_4)$ when $b_2<0$.
Thus, $\|H\|_{\mathcal H_\infty}<1$ holds if and
only if
\begin{equation}
\label{eq:v2v_gain_conditions}
    b_4>0,
    \qquad
    b_0>0,
    \qquad
    b_2>-2\sqrt{b_4b_0}.
\end{equation}
These conditions give $|H(j\omega)|<1$ at every
finite $\omega$, and $b_4>0$ is exactly the
condition $-2<k_{LDD}<0$ found above, which by
$H(\infty)=1+k_{LDD}$ is $|H(\infty)|<1$.
Since $|H(j\omega)|$ is continuous and converges as
$|\omega|\rightarrow\infty$, its supremum is either
attained at a finite frequency or equals
$|H(\infty)|$; both are below one, so
$\|H\|_{\mathcal H_\infty}<1$.

\begin{cor}[Robust $\mathcal L_2$ lateral string stability under LFP]
\label{cor:v2v_robust_l2}
Suppose that $k_1>0$, $k_2>0$, and the learning gains are such that
$\|H\|_{\mathcal H_\infty}<1$. Then the platoon governed by
\eqref{eq:lfp_mu}--\eqref{eq:lfp_update}, with performance output
$y_i=e_{lat,i}^{des}$, is robustly
$\mathcal L_2$ lateral string stable.
\end{cor}
\begin{proof}
Appendix~\ref{app:v2v_realization} constructs a state-space
realization with state $x_i$, output $y_i=e_{lat,i}^{des}$,
and local inputs
\begin{equation}
\label{eq:lfp_psi_map}
    \psi_1:=w_1,
    \qquad
    \psi_i:=w_i-w_{i-1},
    \qquad i=2,\ldots,N.
\end{equation}
Its state matrix is Hurwitz for $k_1,k_2>0$, so
$\|H\|_{\mathcal H_\infty}<1$ allows
Theorem~\ref{thm:general_l2_condition} to give the
$\mathcal L_2$ bound in terms of $x_i(0)$ and $\psi_i$.
The appendix also bounds $x_i(0)$ in terms of $x_{e,i}(0)$
uniformly in $N$, while \eqref{eq:lfp_psi_map} gives
\[
\max_{1\le i\le N}\|\psi_i\|_{\mathcal L_2}
\le2\max_{1\le j\le N}\|w_j\|_{\mathcal L_2}.
\]
These are the two bounds required by
Remark~\ref{rem:psi_w}, which converts the theorem's bound
into Definition~\ref{def:robust_l2_lss}.
\end{proof}

Applying Corollary~\ref{cor:uniform_linf_bound} to the same
realization gives an $\mathcal L_\infty$ bound in terms of
$x_i(0)$ and $\psi_i$. The appendix's initial-condition bound
and the perturbation bound above then express it in terms
of $x_{e,i}(0)$ and $w_i$, with coefficients independent of
platoon length.

\textit{\textbf{Learning gains for downstream error attenuation:}}
The DT baseline also satisfies the robust
$\mathcal L_2$ lateral string stability definition:
setting $v_i\equiv0$ in \eqref{eq:lfp_mu} gives each
vehicle stable, decoupled lateral error dynamics.
This guarantees boundedness but does not provide a
mechanism for reducing lateral errors along the platoon.
We therefore examine how LFP can achieve downstream
error attenuation.

As successive vehicles traverse the same road, they
experience common perturbations, such as those caused
by road bank, at the same desired-path location. Information from the predecessor's traversal
can then be used to reduce the follower's lateral
error. The following gain choice makes it possible
to characterize this attenuation in both the
$\mathcal L_2$ and $\mathcal L_\infty$ norms.

Although $\|H\|_{\mathcal H_\infty}<1$ guarantees
$\mathcal L_2$ contraction of the predecessor-to-follower
map, it does not generally imply
$\mathcal L_\infty$ contraction
\cite{swaroop1994string,desoer2009feedback}.
To obtain contraction in both norms, consider the gain
choice
\begin{equation}
\label{eq:alpha_gains}
    k_{LP}=-\alpha k_1,
    \qquad
    k_{LD}=-\alpha k_2,
    \qquad
    k_{LDD}=-\alpha.
\end{equation}
This gives $K_L(s)=-\alpha D(s)$, so that the transfer
function \eqref{eq:v2v_H} reduces to the constant
\begin{equation}
\label{eq:alpha_H}
    H\equiv1-\alpha.
\end{equation}
For $0<\alpha<2$, \eqref{eq:alpha_H} gives
$\|H\|_{\mathcal H_\infty}=|1-\alpha|<1$, establishing
robust $\mathcal L_2$ lateral string stability by
Corollary~\ref{cor:v2v_robust_l2}.
Moreover, since $H(s)=1-\alpha$ is constant, the
predecessor-to-follower map is strictly contractive
in both the $\mathcal L_2$ and $\mathcal L_\infty$
norms, with induced gain $|1-\alpha|$.

A stronger statement is available when part of the perturbation repeats
across successive traversals of the same location. Decompose
\begin{equation}
\label{eq:perturbation_decomposition}
    w_i(l_d)=w_c(l_d)+\tilde w_i(l_d),
\end{equation}
where $w_c$ is the component common to all vehicles at a given desired-path
location, and $\tilde w_i$ is the vehicle-specific residual. 

At \eqref{eq:alpha_gains} the coefficients on the right of
\eqref{eq:lfp_recursion} are $1-\alpha$ times those on the left, and the
perturbations enter only through $w_i-w_{i-1}$, so the common component
cancels and \eqref{eq:lfp_recursion} rearranges to
\begin{align}
\label{eq:transfer_recursion_residual}
    &\left(e_{lat,i}^{des}-(1-\alpha)e_{lat,i-1}^{des}\right)''
    +k_2\left(e_{lat,i}^{des}-(1-\alpha)e_{lat,i-1}^{des}\right)'
    \notag\\
    &+k_1\left(e_{lat,i}^{des}-(1-\alpha)e_{lat,i-1}^{des}\right)
    =
    \tilde w_i-\tilde w_{i-1},
    \quad i=2,\ldots,N.
\end{align}
\begin{prop}[Attenuation under a common perturbation]
\label{prop:transfer_common}
Let $k_1>0$, $k_2>0$, let the learning gains be \eqref{eq:alpha_gains} with
$0<\alpha<2$, let $\tilde w_i\equiv0$ for all $i$, and let the initial
conditions be zero. Then
\begin{equation}
\label{eq:transfer_common_contraction}
    e_{lat,i}^{des}=(1-\alpha)e_{lat,i-1}^{des},
    \qquad i=2,\ldots,N,
\end{equation}
so that
$\|e_{lat,i}^{des}\|_{\mathcal L_2}=|1-\alpha|\,\|e_{lat,i-1}^{des}\|_{\mathcal L_2}$
and
$\|e_{lat,i}^{des}\|_{\mathcal L_\infty}=|1-\alpha|\,\|e_{lat,i-1}^{des}\|_{\mathcal L_\infty}$,
with $|1-\alpha|<1$.
\end{prop}
\begin{proof}
With $\tilde w_i\equiv0$ the right side of
\eqref{eq:transfer_recursion_residual} vanishes, so
$e_{lat,i}^{des}-(1-\alpha)e_{lat,i-1}^{des}$ satisfies the homogeneous
equation with zero initial conditions and is therefore identically zero.
This is \eqref{eq:transfer_common_contraction}, and both norm identities
follow.
\end{proof}
When the perturbations differ among vehicles, the
following proposition bounds the lateral error norms
by envelopes determined by the leader's error and
the differences between consecutive residuals.
\begin{prop}[Downstream envelope with vehicle-specific residuals]
\label{prop:transfer_envelope}
Let $N\ge2$, $k_1>0$, $k_2>0$, let the learning gains be \eqref{eq:alpha_gains} with
$0<\alpha\le1$, let the initial conditions be zero, and let
\begin{equation}
\label{eq:mismatch_parameter}
    \Delta_{\tilde w}
    :=
    \max_{2\le i\le N}
    \left\|
        \tilde w_i-\tilde w_{i-1}
    \right\|_{\mathcal L_2}.
\end{equation}
Then, with
$B_2:=\|T_{y\psi}\|_{\mathcal H_\infty}\Delta_{\tilde w}/\alpha$, for $i=2,\ldots,N$,
\begin{equation}
\label{eq:transfer_l2_envelope}
    \left\|e_{lat,i}^{des}\right\|_{\mathcal L_2}
    \le
    B_2
    +
    \left(1-\alpha\right)^{i-1}
    \left(
        \left\|e_{lat,1}^{des}\right\|_{\mathcal L_2}-B_2
    \right),
\end{equation}
and, with
$B_\infty:=\|h_{y\psi}\|_{\mathcal L_2}\Delta_{\tilde w}/\alpha$ and
$h_{y\psi}$ as in \eqref{eq:general_y_impulse_responses}, for $i=2,\ldots,N$,
\begin{equation}
\label{eq:transfer_linf_envelope}
    \left\|e_{lat,i}^{des}\right\|_{\mathcal L_\infty}
    \le
    B_\infty
    +
    \left(1-\alpha\right)^{i-1}
    \left(
        \left\|e_{lat,1}^{des}\right\|_{\mathcal L_\infty}-B_\infty
    \right).
\end{equation}
\end{prop}
\begin{proof}
Applying $\|T_{y\psi}\|_{\mathcal H_\infty}$ as the induced $\mathcal L_2$
gain to \eqref{eq:transfer_recursion_residual} and using the triangle
inequality gives
$\|e_{lat,i}^{des}\|_{\mathcal L_2}\le
(1-\alpha)\|e_{lat,i-1}^{des}\|_{\mathcal L_2}
+\|T_{y\psi}\|_{\mathcal H_\infty}\Delta_{\tilde w}$, where
$1-\alpha\ge0$ because $\alpha\le1$. Iterating this recursion from the
leader gives, for $i=2,\ldots,N$,
$\|e_{lat,i}^{des}\|_{\mathcal L_2}\le
\left(1-\alpha\right)^{i-1}\|e_{lat,1}^{des}\|_{\mathcal L_2}
+\|T_{y\psi}\|_{\mathcal H_\infty}\Delta_{\tilde w}
\sum_{j=0}^{i-2}\left(1-\alpha\right)^{j}$, and
$\sum_{j=0}^{i-2}(1-\alpha)^{j}
=\left[1-\left(1-\alpha\right)^{i-1}\right]/\alpha$, so the second term
is $B_2\left[1-\left(1-\alpha\right)^{i-1}\right]$. Rearranging gives
\eqref{eq:transfer_l2_envelope}. Equation
\eqref{eq:transfer_linf_envelope} follows identically, bounding the
convolution by
$\|h_{y\psi}\|_{\mathcal L_2}\|\tilde w_i-\tilde w_{i-1}\|_{\mathcal L_2}$ as in
the proof of Corollary~\ref{cor:uniform_linf_bound}.
\end{proof}
Both envelopes have the form
$B+(1-\alpha)^{i-1}(\|e_{lat,1}^{des}\|-B)$.
For $0<\alpha\le1$, each envelope approaches $B$,
decreasing when the leader's norm exceeds $B$ and
increasing when it is below $B$. In the former case,
the bound for every follower lies below the leader's
norm, although the actual norms need not decrease
monotonically between consecutive vehicles.

When all vehicles experience the same perturbation,
both floors vanish and
Proposition~\ref{prop:transfer_common} gives exact
contraction. Otherwise, the floors are proportional
to $\Delta_{\tilde w}$, which measures perturbation
differences between consecutive vehicles.
The common component $w_c$ affects the leader's
error but cancels from these differences and
therefore does not affect the floors.

These results show how vehicle-level control and
LFP can complement each other. Robust or adaptive
control can reduce each vehicle's response to local
perturbations, while LFP uses predecessor information
to attenuate lateral errors caused by perturbations
that repeat along the platoon. Differences between
vehicles' perturbations enter the error bounds
through $\Delta_{\tilde w}$ and set the floors
of the derived envelopes.

    \section{Numerical Evaluation}
\label{sec:numerical_evaluation}
This section compares PT, the DT baseline, and LFP
introduced in Section~\ref{sec:platoon_control}.
The comparison distinguishes the benefit of sharing
the desired path from the additional error attenuation
obtained through the LFP strategy.

Subsection~\ref{subsec:experiment_reduced} evaluates
the reduced arc-length error dynamics with prescribed
perturbations, examining the effects of perturbation
repeatability and platoon length.
Subsection~\ref{subsec:experiment_physical} evaluates
the three configurations using time-domain vehicle
dynamics under heterogeneity, parameter uncertainty,
varying longitudinal speed, and road bank.

The three configurations are as follows. PT uses \eqref{eq:pt_mu} with
\eqref{eq:pt_error_relation}, DT uses
\eqref{eq:lfp_mu} with $v_i\equiv0$, and LFP uses
\eqref{eq:lfp_mu}--\eqref{eq:lfp_update} with
the learning gains \eqref{eq:alpha_gains}.
The local feedback gains are fixed at $k_1=0.04~{\rm m}^{-2}$ and
$k_2=0.28~{\rm m}^{-1}$, and $\alpha=0.47$ throughout, so that $H\equiv0.53$ by
\eqref{eq:alpha_H}. The constants entering
Proposition~\ref{prop:transfer_envelope} are then $\|T_{y\psi}\|_{\mathcal
H_\infty}=25.005~{\rm m}^2$ and $\|h_{y\psi}\|_{\mathcal L_2}=6.682~{\rm
m}^{3/2}$.

All norms are taken with respect to the desired-path arc length $l_d$.
Definition~\ref{def:robust_l2_lss} and Theorem~\ref{thm:general_l2_condition}
are stated over $l_d\in[0,\infty)$, whereas a driving operation covers a
finite path and the error dynamics have a decaying tail after the
perturbations cease. In the reduced-model evaluation the perturbations are
therefore set to zero beyond the operating path and the integration is
continued over a tail long enough that the free-response envelope of every
vehicle ends below $10^{-6}$ of its peak, so
that the reported norms approximate the infinite-horizon quantities. In the
heterogeneous-vehicle evaluation the norms are
reported over the path actually traveled; they measure finite-operation
performance and are not used to verify the bounds.

\subsection{Reduced-model evaluation}
\label{subsec:experiment_reduced}
The error dynamics of the three configurations are integrated over a
$160$~m path. The first
study isolates the effect of repeatability at a fixed platoon length, and the
second varies the platoon length under a fixed perturbation construction.

The first study examines how the downstream attenuation depends on the
repeatability of the perturbation. Let $\bar w>0$ be a fixed perturbation
level and let $\chi_c,\chi_1,\ldots,\chi_N$ be fixed spatial profiles with
$\|\chi_c\|_{\mathcal L_2}=\|\chi_i\|_{\mathcal L_2}=1$ and
$\langle\chi_c,\chi_i\rangle=0$, where $\langle\cdot,\cdot\rangle$ is the
$\mathcal L_2$ inner product over the operating path. Each is generated once
as a finite sum of spatial sinusoids with randomly drawn spatial frequencies
and phases, tapered smoothly to zero at both ends of the operating path, so that the
zero extension beyond it is continuous, and
the $\chi_i$ are then orthogonalized against $\chi_c$ by Gram--Schmidt and
renormalized. For a scalar $\beta\in[0,\bar w]$, let
\begin{equation}
\label{eq:test2_decomposition}
    w_i
    =
    \sqrt{\bar w^2-\beta^2}\,\chi_c
    +
    \beta\chi_i,
    \qquad i=1,\ldots,N,
\end{equation}
so that the two components of \eqref{eq:perturbation_decomposition} are
$w_c=\sqrt{\bar w^2-\beta^2}\,\chi_c$ and $\tilde w_i=\beta\chi_i$. Since
$\chi_c$ and $\chi_i$ are orthogonal the cross term vanishes, and
$\|w_i\|_{\mathcal L_2}^2 =(\bar w^2-\beta^2)\|\chi_c\|_{\mathcal L_2}^2
+\beta^2\|\chi_i\|_{\mathcal L_2}^2=\bar w^2$ for every vehicle and every
$\beta$. Increasing $\beta$ therefore transfers perturbation energy from the
common component to the vehicle-specific one while leaving the magnitude of
the perturbation acting on each vehicle unchanged, which separates
repeatability from perturbation magnitude. With $\tilde w_i=\beta\chi_i$ the quantity
$\Delta_{\tilde w}$ of \eqref{eq:mismatch_parameter} is
$\beta\max_{2\le i\le N}\|\chi_i-\chi_{i-1}\|_{\mathcal L_2}$, so it is
proportional to $\beta$, and the envelopes below use its exact value
computed from the generated profiles. The first study uses $N=10$, $\bar w=0.07~{\rm m}^{-1/2}$, and zero initial
conditions, as required by Proposition~\ref{prop:transfer_envelope}.

\begin{figure}[h]
    \centering
    \subfloat[PT]{%
        \includegraphics[width=0.48\columnwidth]
        {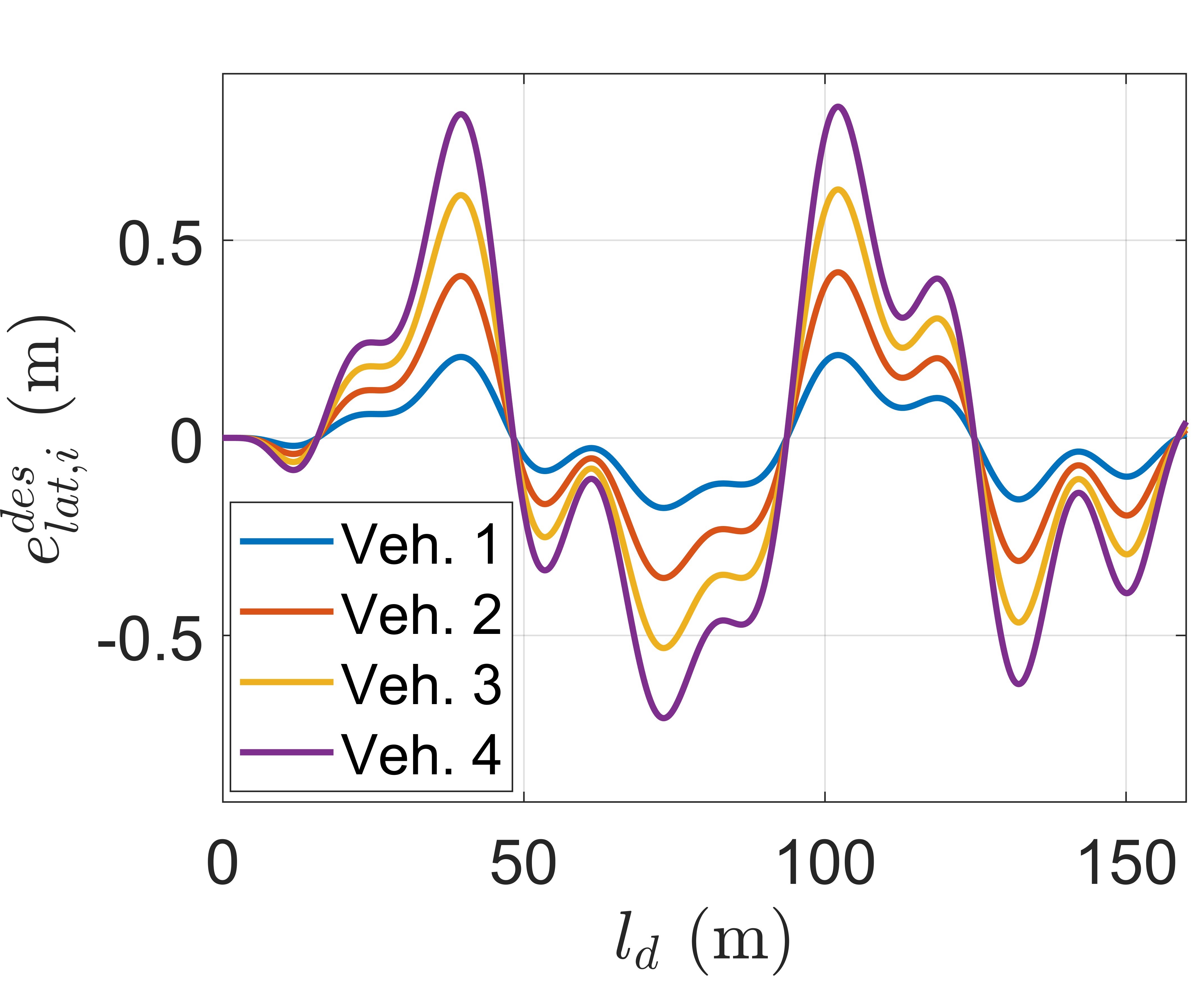}%
    }
    \hfil
    \subfloat[DT]{%
        \includegraphics[width=0.48\columnwidth]
        {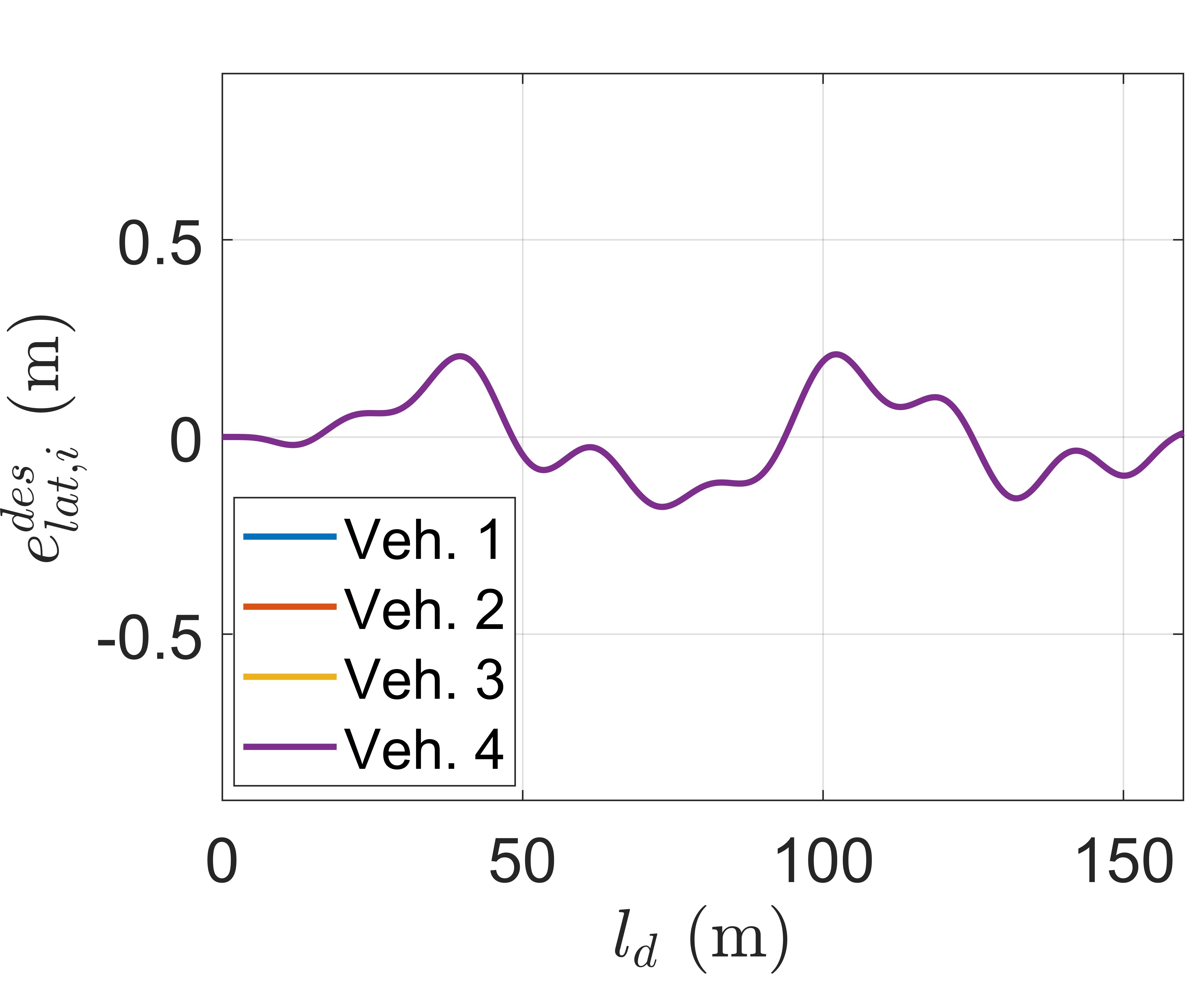}%
    }

    \subfloat[LFP]{%
        \includegraphics[width=0.48\columnwidth]
        {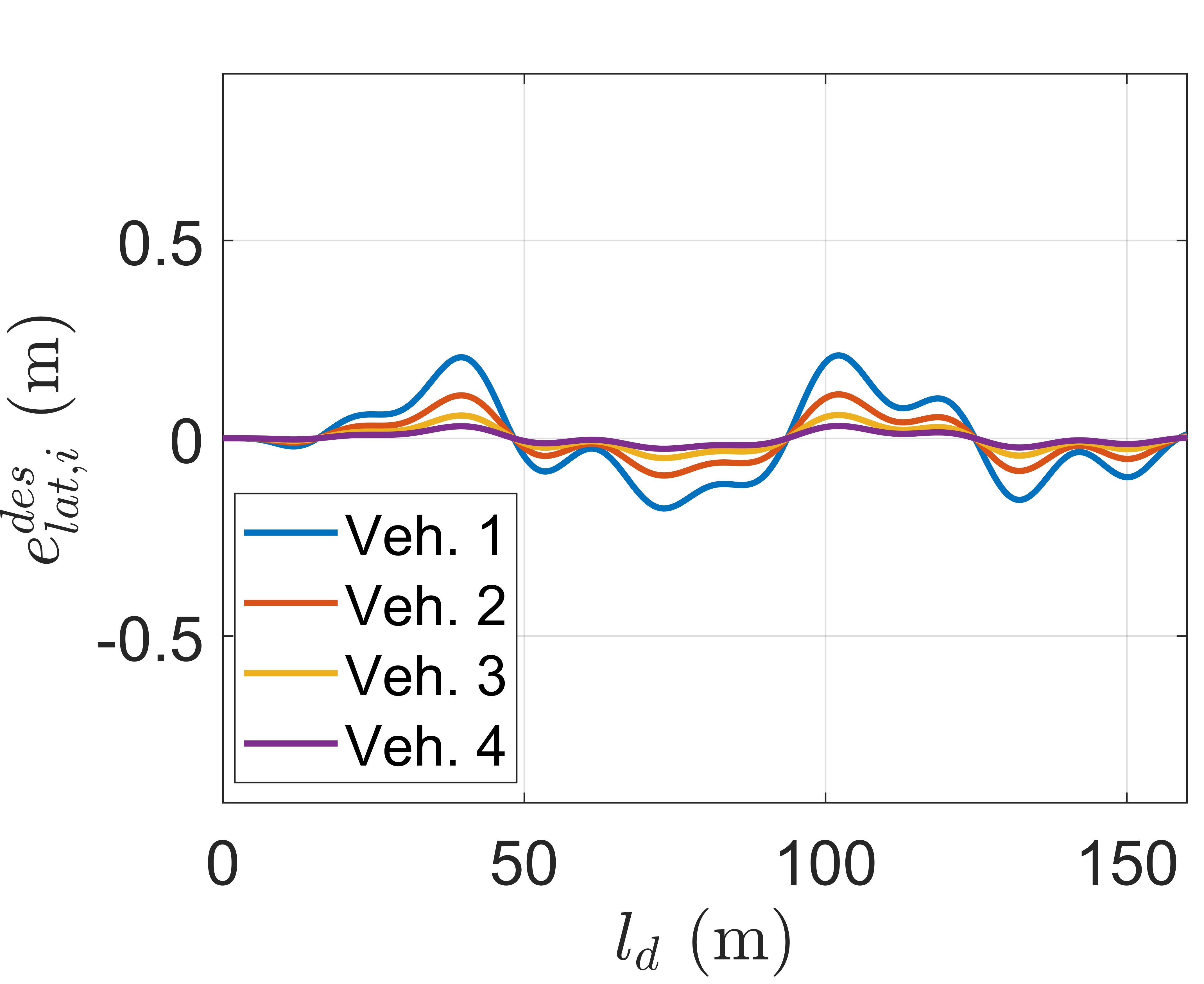}%
    }
    \caption{Desired-path lateral errors of vehicles $1$--$4$
    under a common perturbation. All panels share a vertical scale.}
    \label{fig:reduced_traces}
\end{figure}

Figure~\ref{fig:reduced_traces} shows the desired-path
lateral errors of vehicles $1$ to $4$ at $\beta=0$,
where every vehicle experiences the same perturbation.
In this case, the analysis gives exact relationships
between the vehicle errors. Under PT the error of
vehicle $i$ is $i$ times the leader's by
\eqref{eq:pt_linear_error_accumulation}. Under DT the four traces are
identical, since the vehicles are decoupled. Under LFP the error of each vehicle is $1-\alpha$
times its predecessor's by
Proposition~\ref{prop:transfer_common}, so the error
of vehicle $4$ is $0.53^3\approx0.15$ times the
leader's error. These responses illustrate the three behaviors
distinguished by the analysis: accumulation,
boundedness without attenuation, and attenuation.

\begin{figure}[h]
    \centering
    \subfloat[$\mathcal L_2$ norm versus vehicle index]{%
        \includegraphics[width=0.48\columnwidth]
        {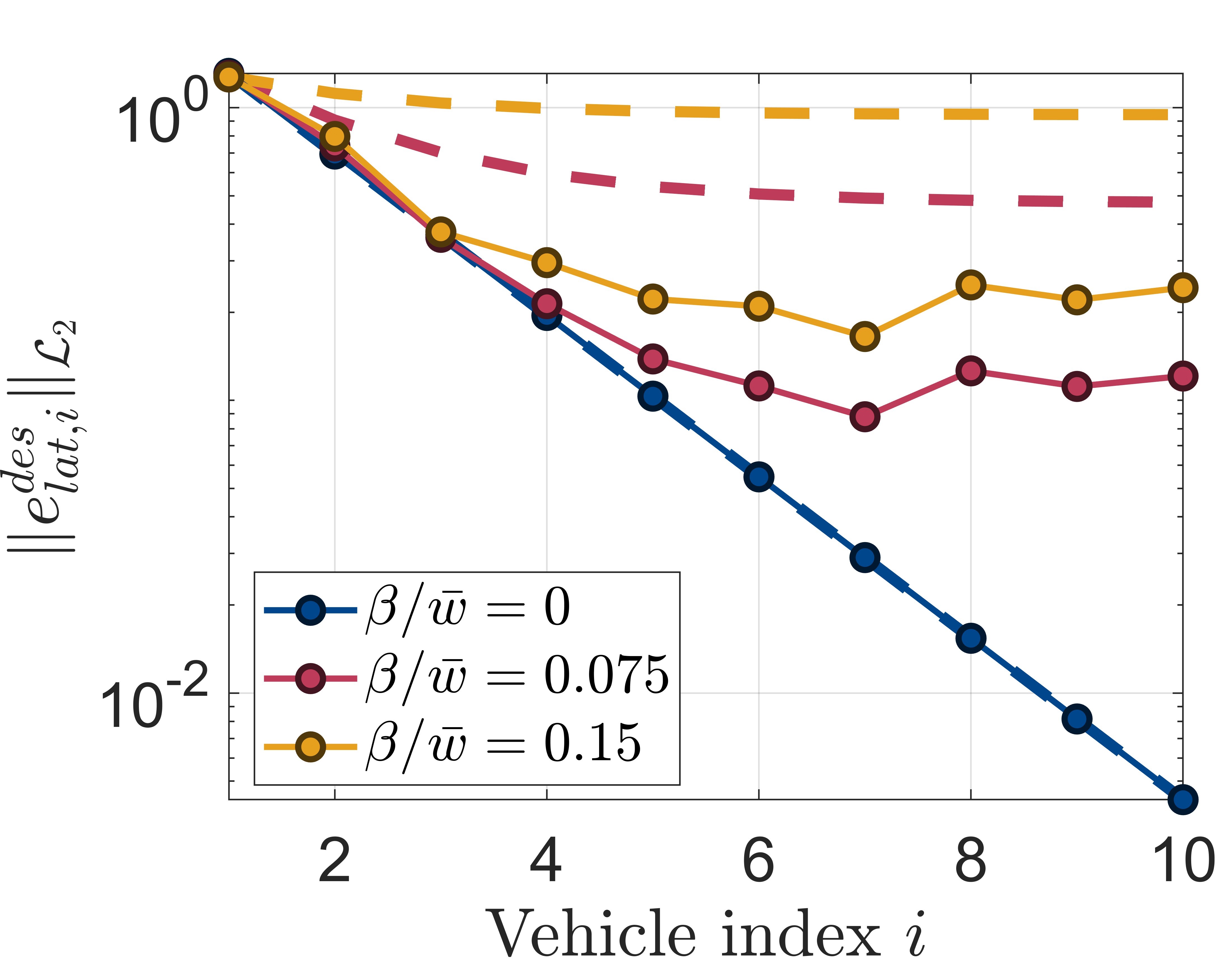}%
    }
    \hfil
    \subfloat[$\mathcal L_\infty$ norm versus vehicle index]{%
        \includegraphics[width=0.48\columnwidth]
        {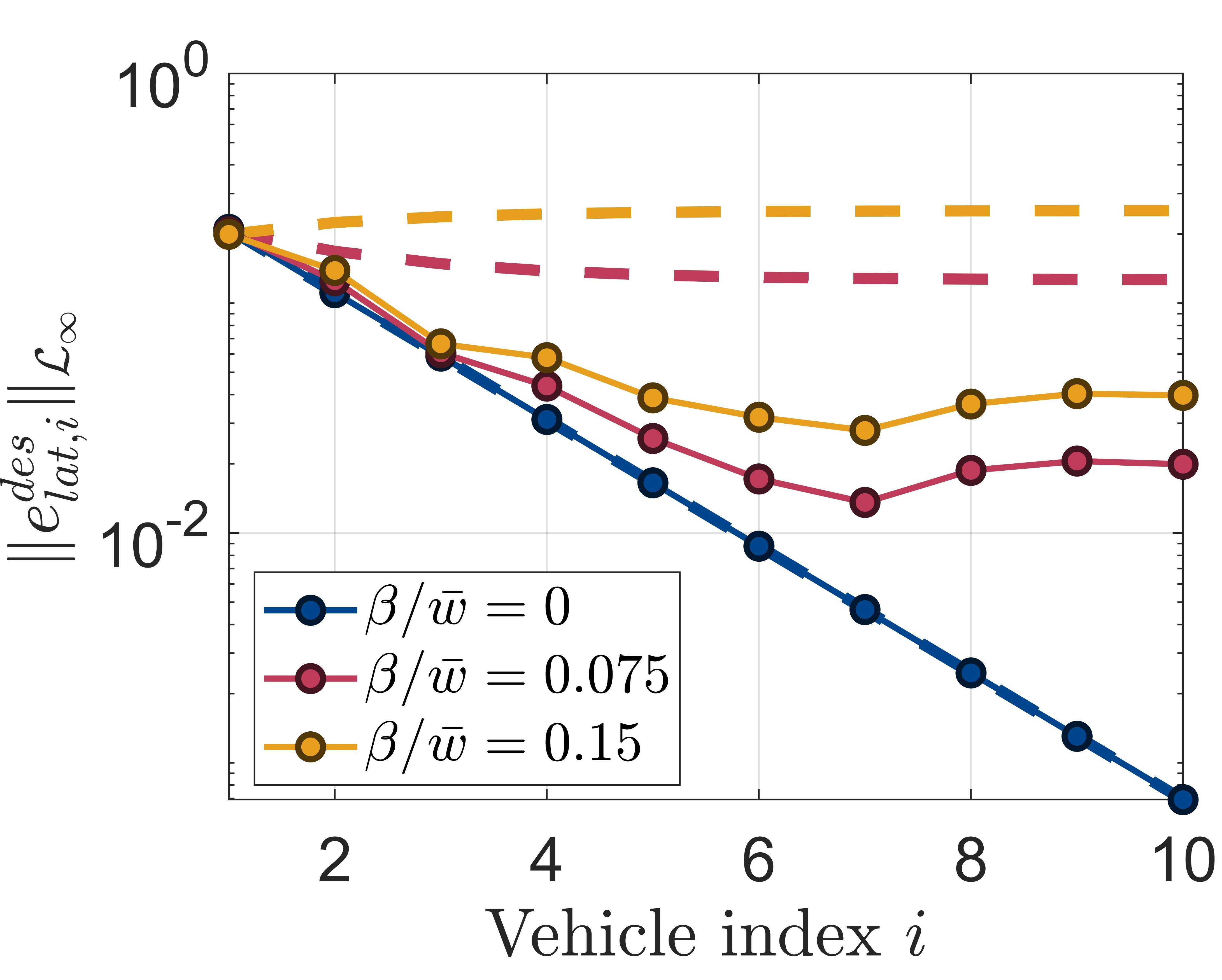}%
    }

    \subfloat[Last-vehicle $\mathcal L_2$ norm]{%
        \includegraphics[width=0.48\columnwidth]
        {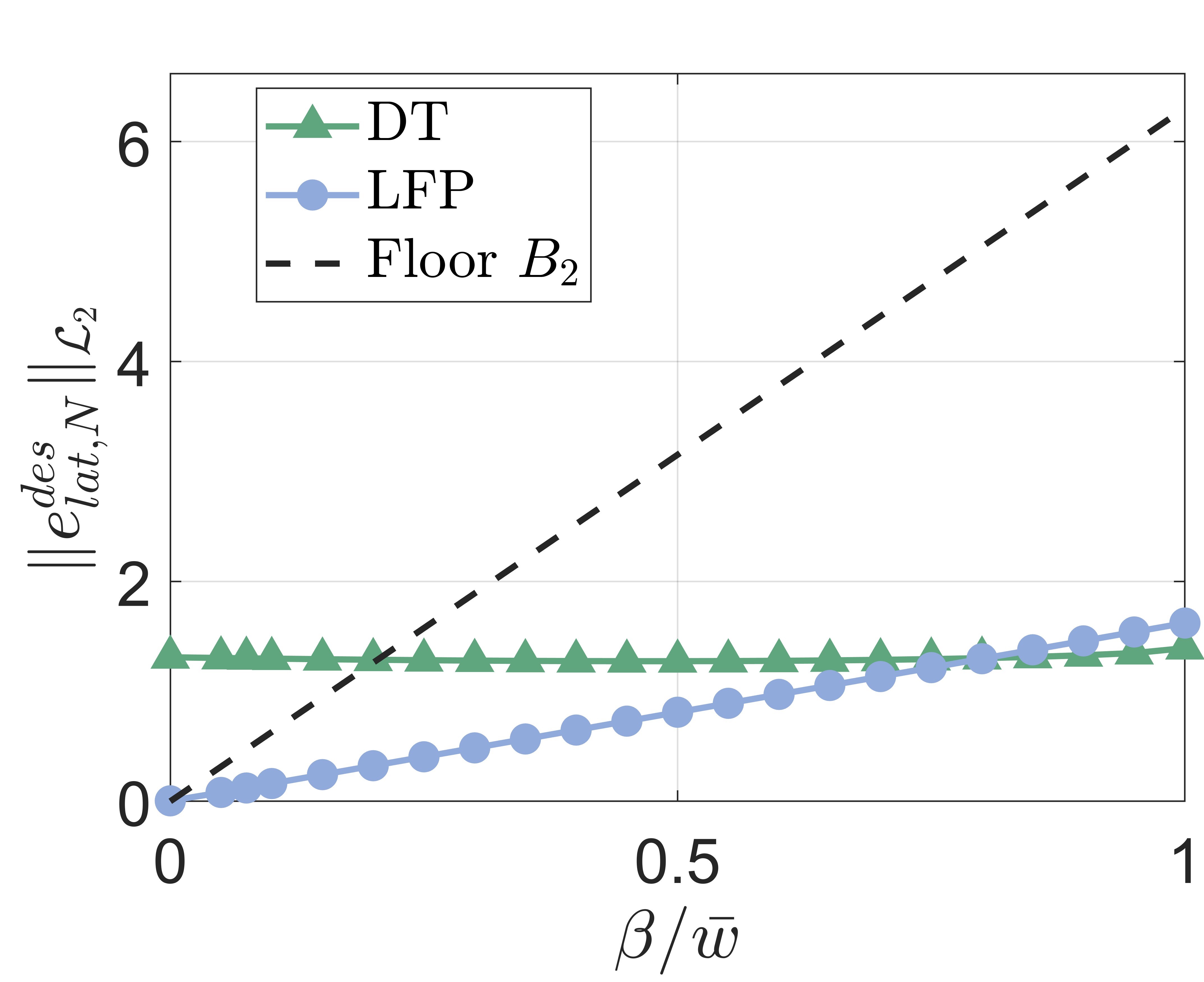}%
    }
    \hfil
    \subfloat[Last-vehicle $\mathcal L_\infty$ norm]{%
        \includegraphics[width=0.48\columnwidth]
        {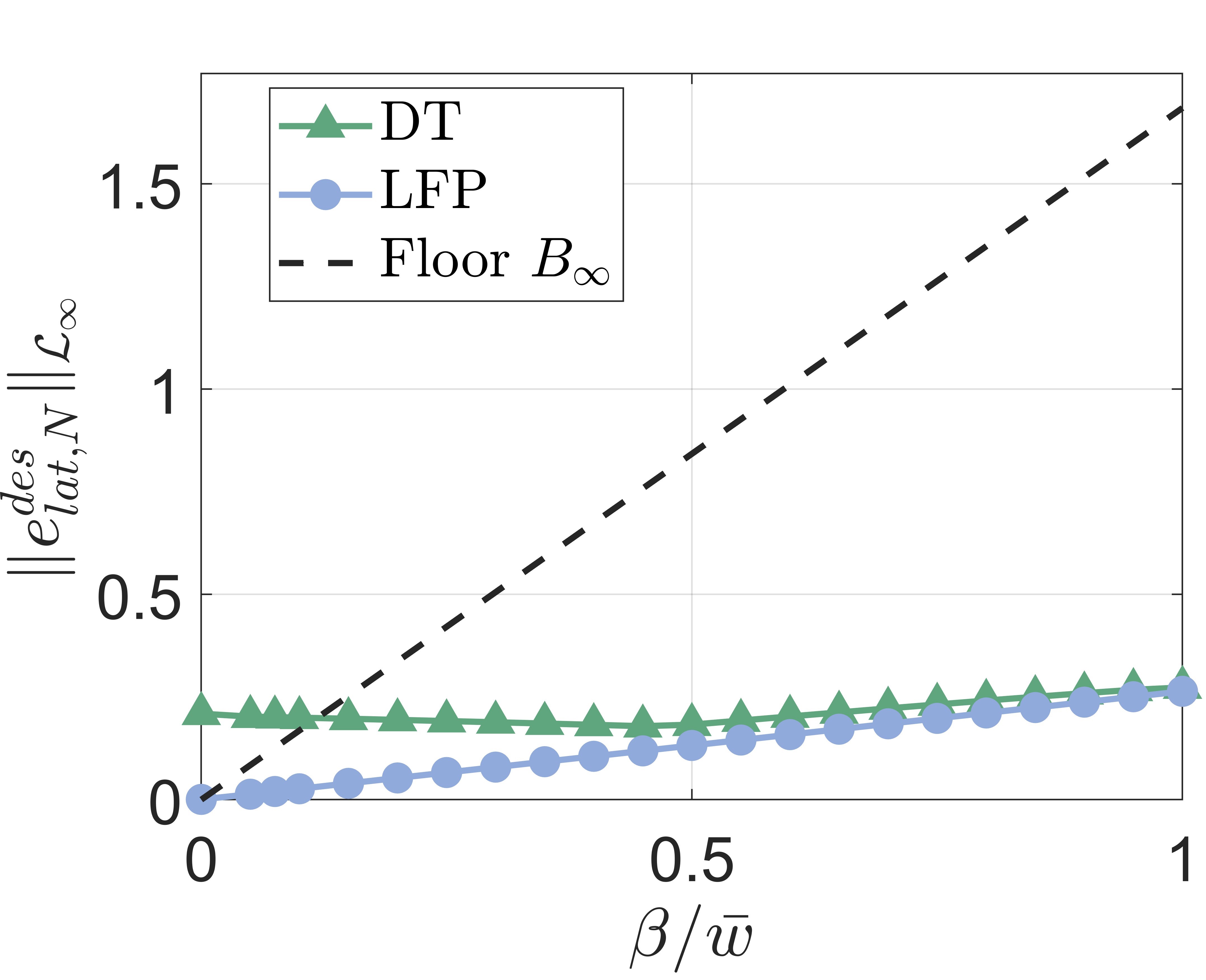}%
    }
    \caption{Effect of perturbation repeatability in the reduced
    model. In (a) and (b), dashed curves show the theoretical
    envelopes; (c) and (d) vary $\beta/\bar w$.}
    \label{fig:reduced_repeatability}
\end{figure}

Figure~\ref{fig:reduced_repeatability}(a) and (b)
show the LFP error norms against vehicle index at
$\beta=0$, $\beta=0.075\bar w$, and
$\beta=0.15\bar w$. Dashed curves show the corresponding
envelopes \eqref{eq:transfer_l2_envelope} and
\eqref{eq:transfer_linf_envelope}.
At $\beta=0$, both floors vanish and the norms
coincide with their envelopes. At the two nonzero
levels, the norms initially decrease and then
fluctuate at levels that rise with $\beta$, while
remaining below their respective envelopes.

Figure~\ref{fig:reduced_repeatability}(c) and (d) show
the last-vehicle norms against $\beta/\bar w$ for DT
and LFP. For the profiles used here, the DT response
is nearly flat, whereas the LFP response rises nearly
linearly as the vehicle-specific component increases.
In the $\mathcal L_2$ norm, DT and LFP meet at
$\beta/\bar w=0.804$; in the $\mathcal L_\infty$ norm,
LFP remains below DT throughout the sweep.
The increasing LFP response is consistent with
Proposition~\ref{prop:transfer_envelope}, whose envelope
floors are proportional to $\Delta_{\tilde w}$.

\begin{figure}[h]
    \centering
    \subfloat[PT]{%
        \includegraphics[width=0.48\columnwidth]
        {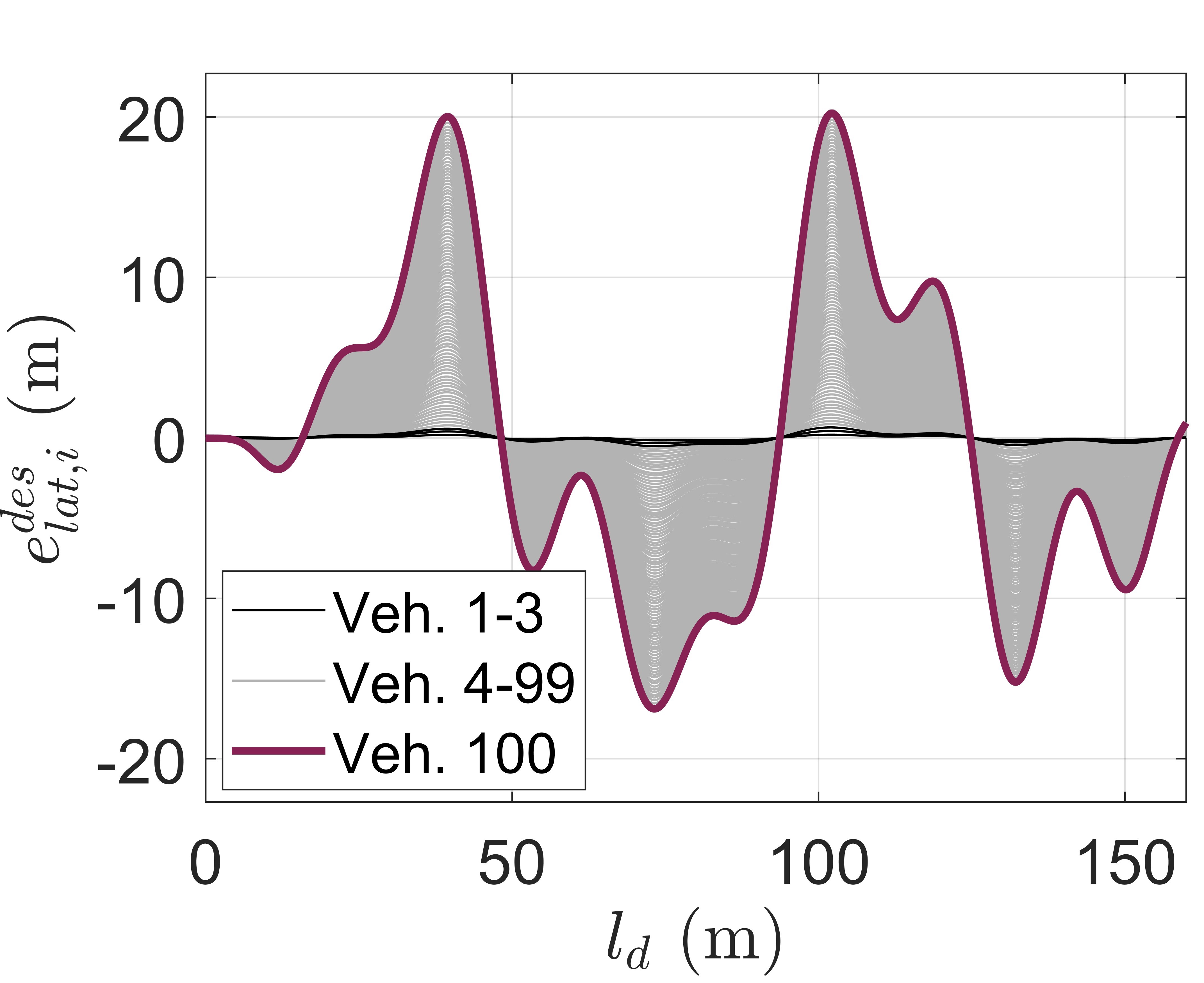}%
    }
    \hfil
    \subfloat[DT]{%
        \includegraphics[width=0.48\columnwidth]
        {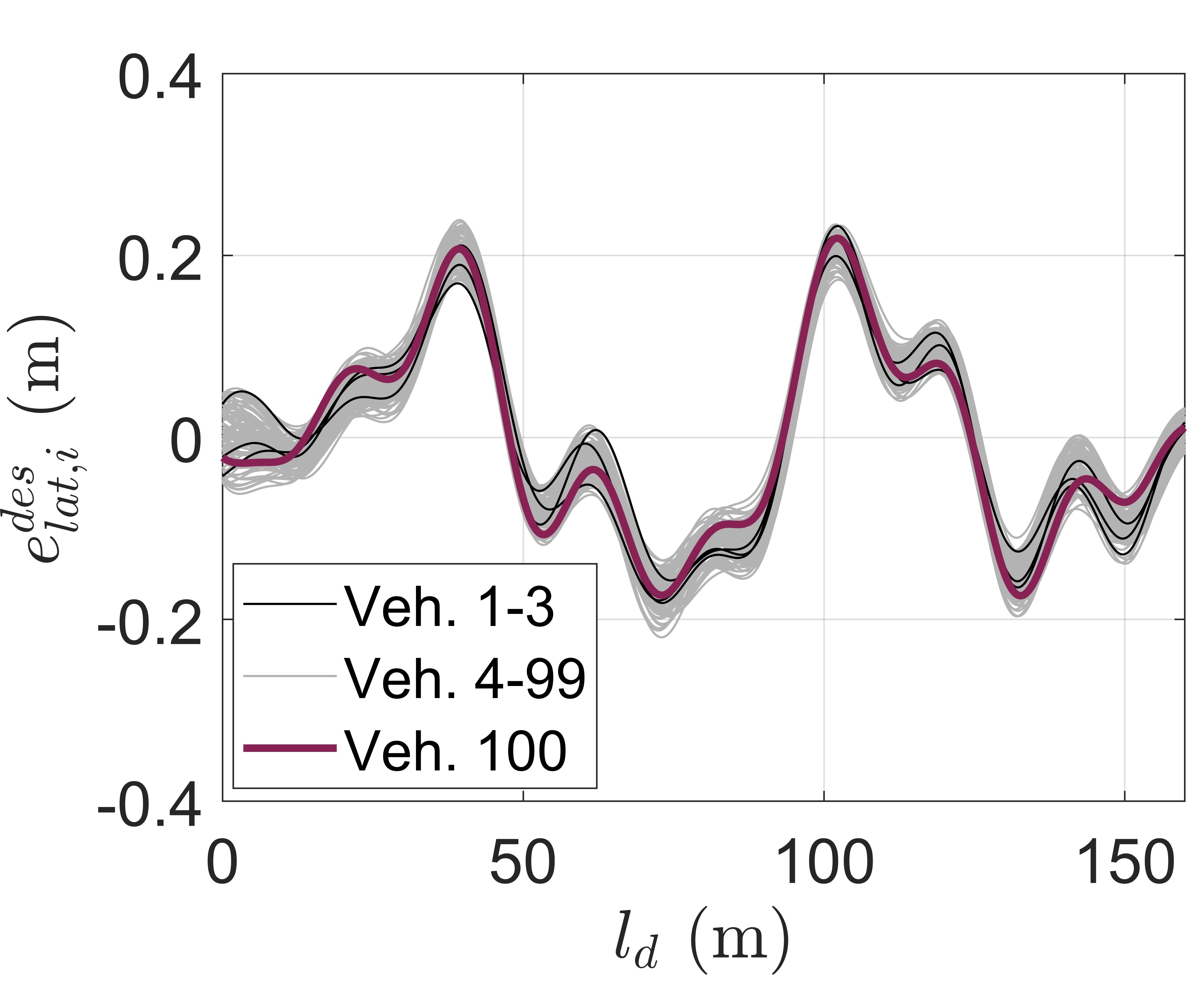}%
    }

    \subfloat[LFP]{%
        \includegraphics[width=0.48\columnwidth]
        {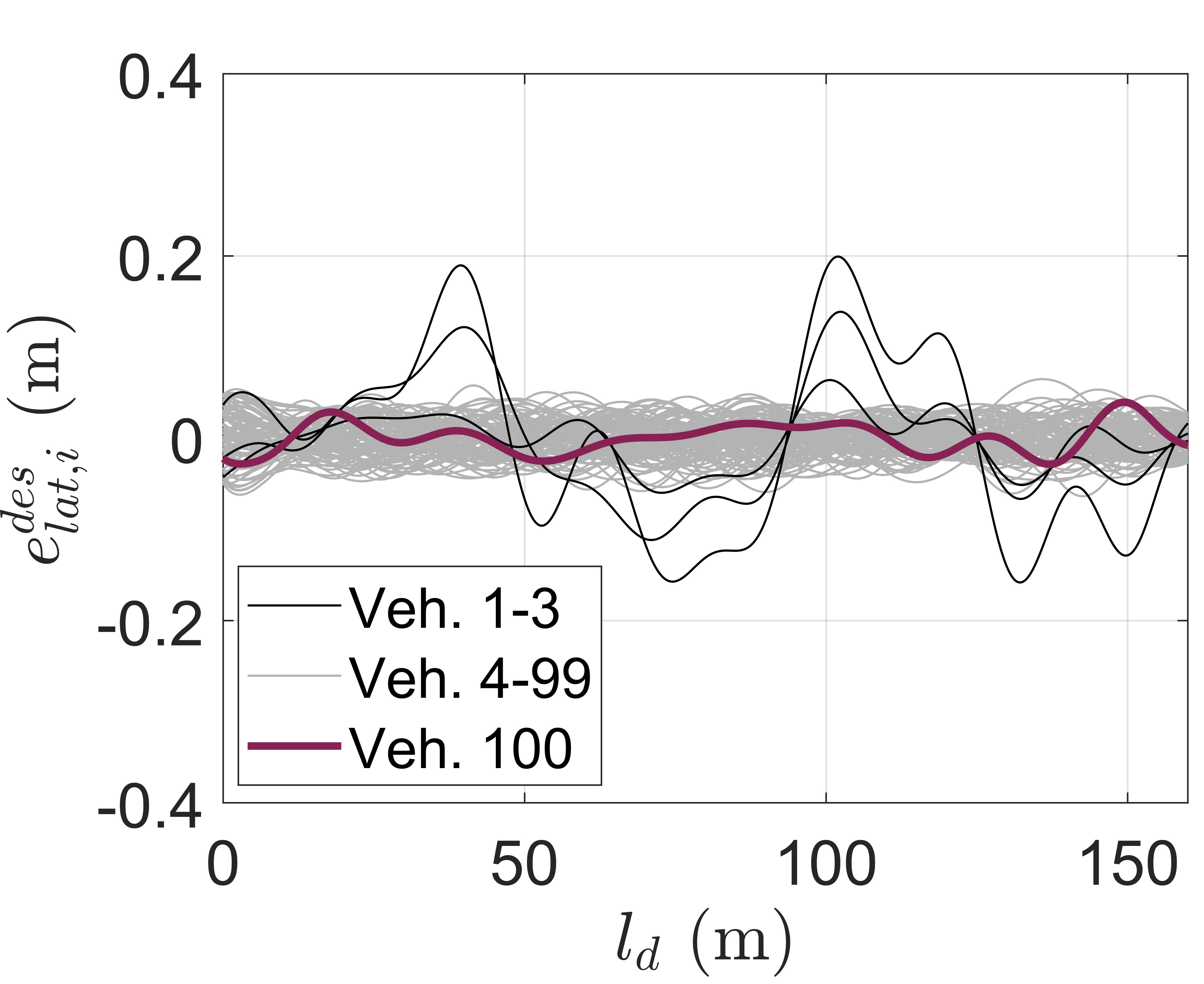}%
    }
    \hfil
    \subfloat[Maximum $\mathcal L_2$ norm versus $N$]{%
        \includegraphics[width=0.48\columnwidth]
        {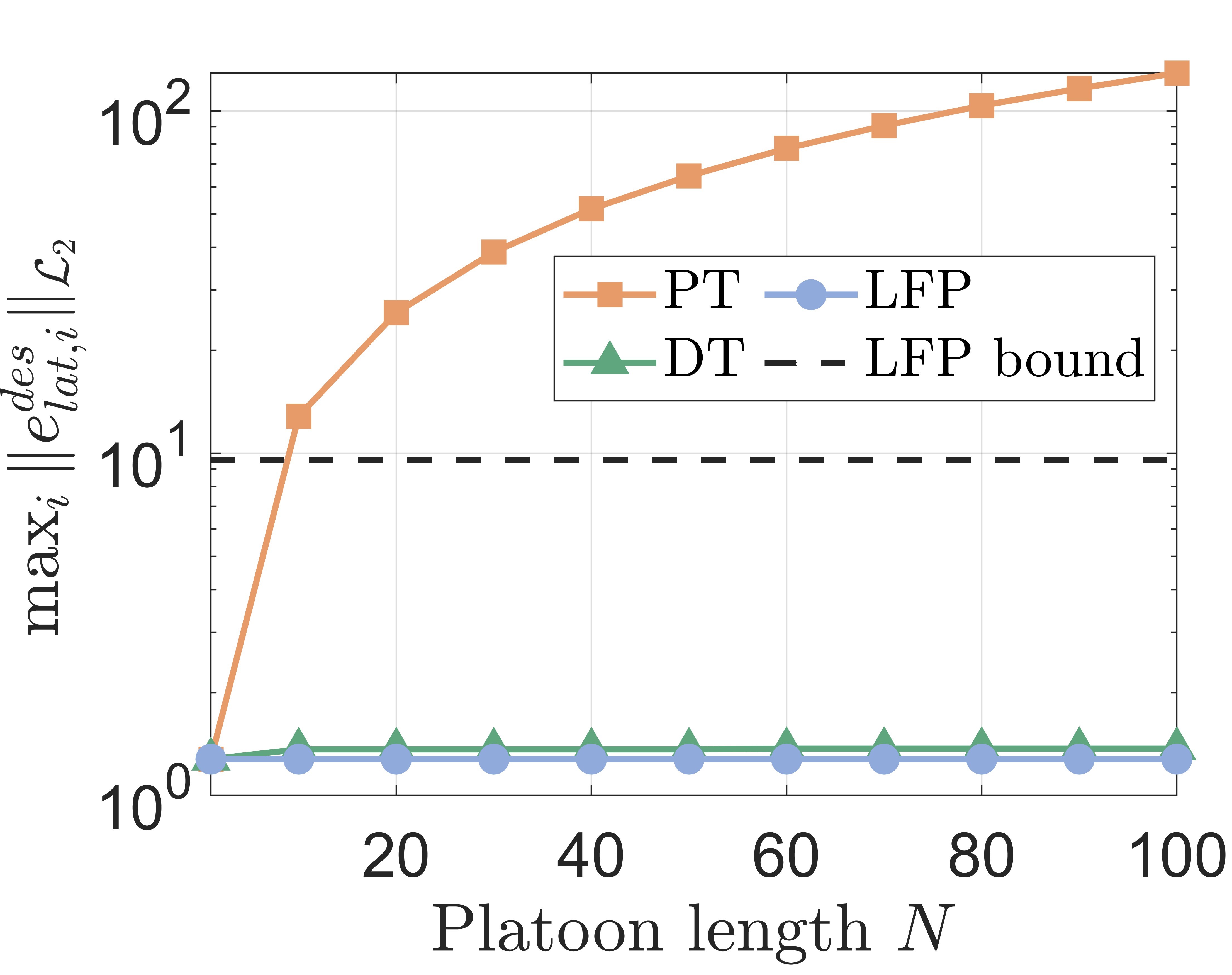}%
    }
    \caption{Reduced-model evaluation with $100$ vehicles:
    (a)--(c) desired-path lateral errors; (d) maximum error
    norm among the first $N$ vehicles.}
    \label{fig:reduced_scaling}
\end{figure}

The second study examines the dependence of the maximum
lateral error norms on platoon length and compares
the LFP responses with the theoretical bounds. It
uses the same construction \eqref{eq:test2_decomposition} at the fixed value
$\beta=0.15\bar w$, now with nonzero initial conditions satisfying
$|e_{lat,i}^{des}(0)|\le0.05$~m and $|(e_{lat,i}^{des})'(0)|\le0.01$, over a
platoon of $100$ vehicles. Figure~\ref{fig:reduced_scaling}(a) to (c) show the error of every vehicle.
Under PT the error grows with the vehicle index, vehicle $100$ reaching
$20.6$~m against the leader's $0.199$~m. Under DT and LFP the errors stay
within a band that does not widen, and under LFP every vehicle past the third lies inside a band well
below the first three.

The bounds of Theorem~\ref{thm:general_l2_condition} and
Corollary~\ref{cor:uniform_linf_bound} are evaluated for LFP from $\bar
X=\max_i\|x_i(0)\|_2$ and $\bar\Psi=\max_i\|\psi_i\|_{\mathcal L_2}
=\max\{\|w_1\|_{\mathcal L_2}, \max_{2\le
i\le100}\|w_i-w_{i-1}\|_{\mathcal L_2}\}$, where $x_i$ and $\psi_i$ denote the
state and local input of the realization in Appendix~\ref{app:v2v_realization}.
$M=\sqrt{\lambda_{\max}(P)/\lambda_{\min}(P)}$ and
$\lambda=1/(2\lambda_{\max}(P))$ are chosen in \eqref{eq:general_exponential_bound}, with $P\succ0$ obtained from $A^TP+PA=-I$. Panel (d) gives the $\mathcal L_2$ maximum over the first $N$ vehicles against
$N$. The PT maximum grows with the platoon length, from $1.278~{\rm m}^{3/2}$ at $N=1$ to
$129.1~{\rm m}^{3/2}$ at $N=100$, while the DT and LFP maxima change little as
$N$ increases. The LFP
maximum is $1.278~{\rm m}^{3/2}$ at every $N$, attained at the leader and
never exceeded by a follower. The DT maximum rises to $1.371~{\rm m}^{3/2}$ as
additional vehicles with different perturbations
and initial conditions are included. For LFP the $\mathcal{L}_2$ bound of
Theorem~\ref{thm:general_l2_condition} is $9.57~{\rm m}^{3/2}$ against the
realized $1.278~{\rm m}^{3/2}$. The bound is evaluated using the initial-condition and
input maxima over all $100$ vehicles, so the same value
bounds the maximum error norm among the first $N$
vehicles for every $1\le N\le100$.
Corollary~\ref{cor:uniform_linf_bound} gives a $1.82$~m $\mathcal{L}_\infty$ bound against a realized
$0.1993$~m maximum $\mathcal L_\infty$ norm, which is not shown in the figure.

\subsection{Heterogeneous-vehicle evaluation}
\label{subsec:experiment_physical}
The three configurations are evaluated using the
time-domain bicycle model in \cite{ch3preprint},
augmented with global-position kinematics.
This study examines lateral error propagation under
vehicle heterogeneity, parameter uncertainty,
varying longitudinal speed, and road bank.

\begin{figure}[h]
    \centering
    \subfloat[Desired path]{%
        \includegraphics[width=0.48\columnwidth]
        {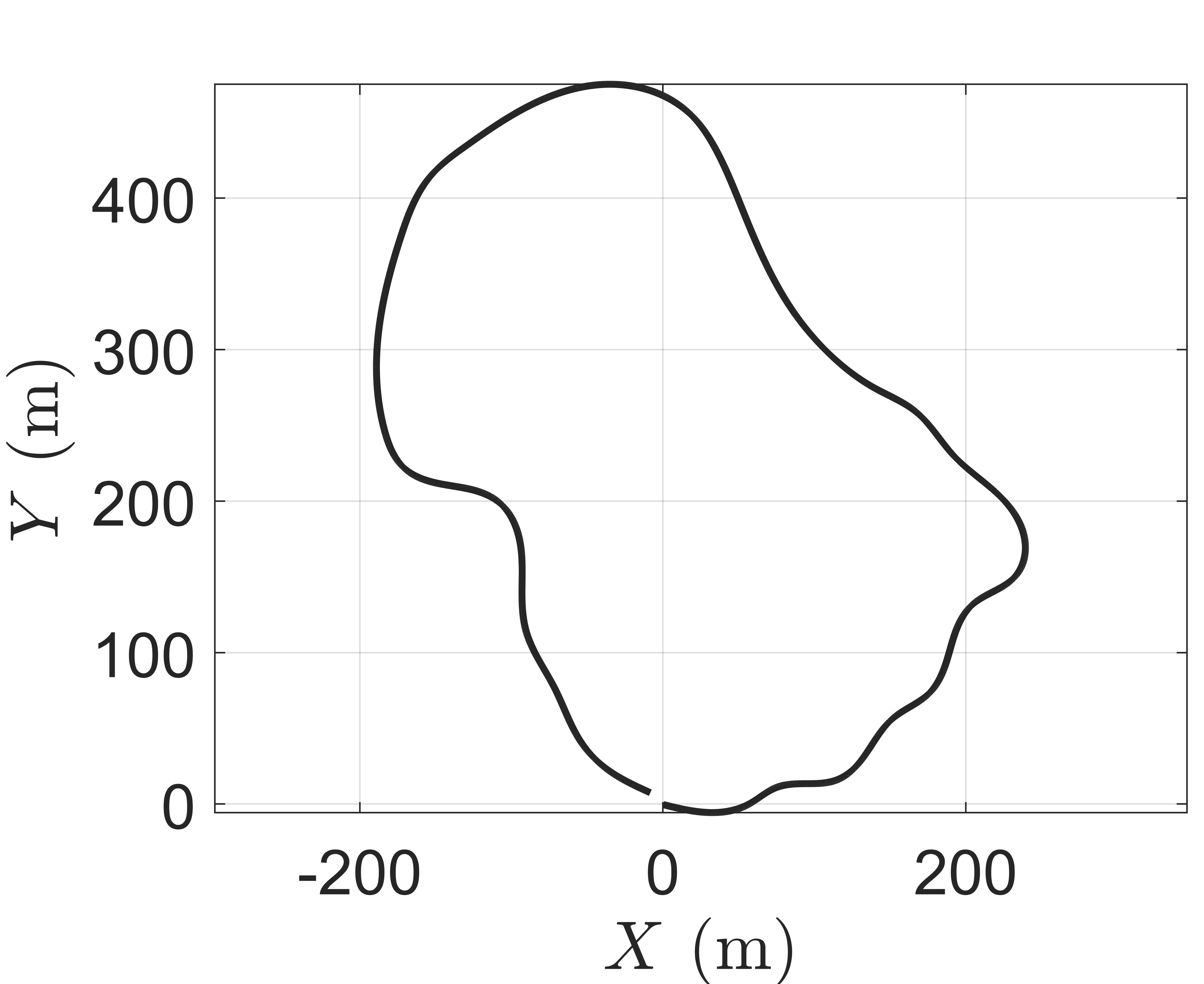}%
    }
    \hfil
    \subfloat[Path curvature]{%
        \includegraphics[width=0.48\columnwidth]
        {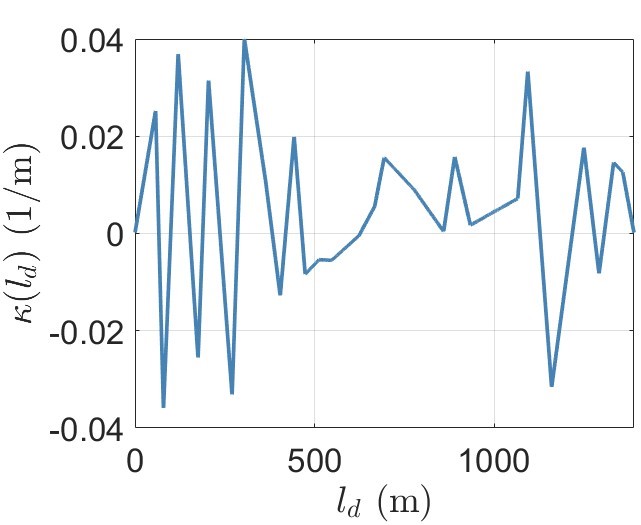}%
    }

    \subfloat[Longitudinal speed]{%
        \includegraphics[width=0.48\columnwidth]
        {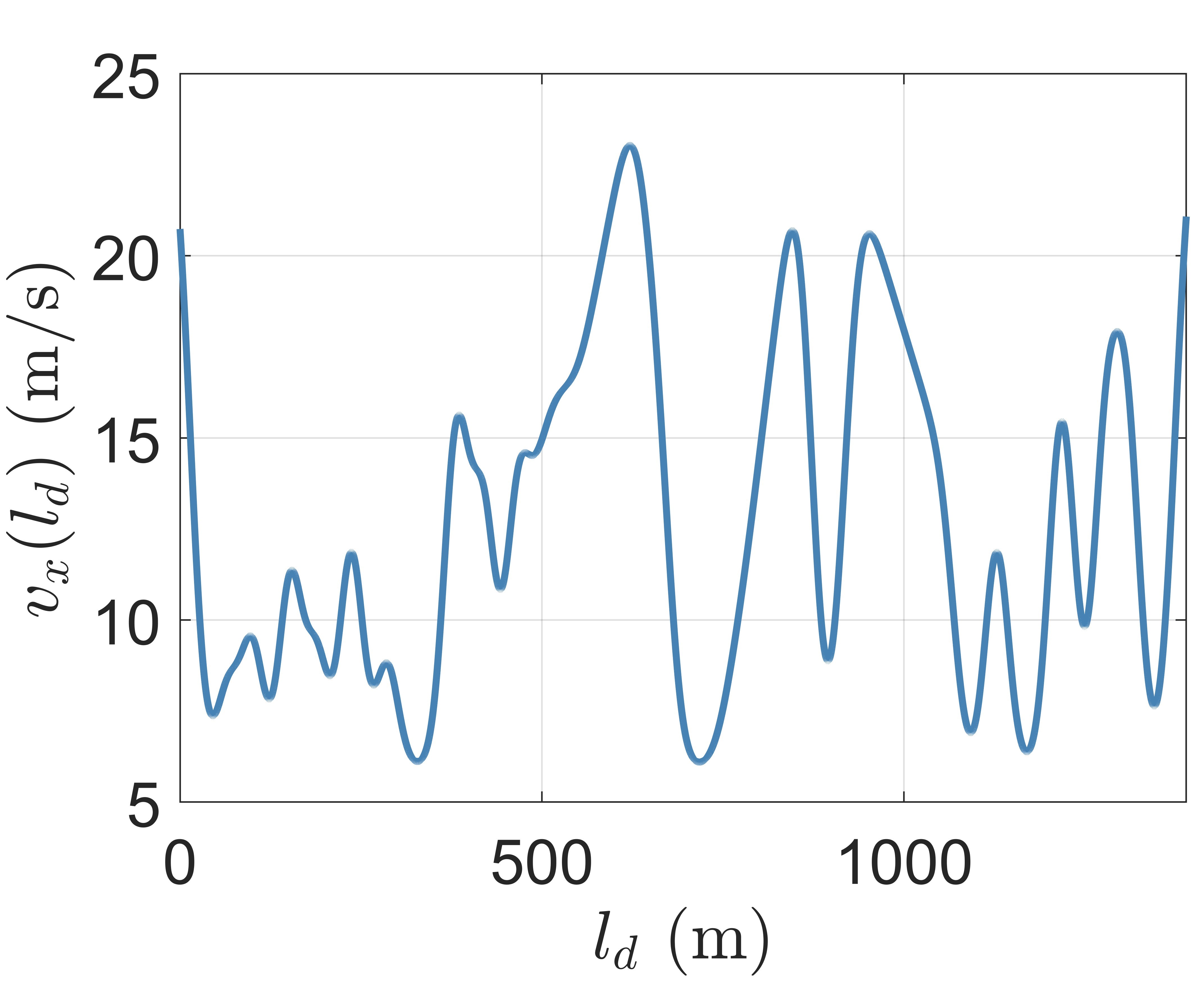}%
    }
    \hfil
    \subfloat[Road bank angle]{%
        \includegraphics[width=0.48\columnwidth]
        {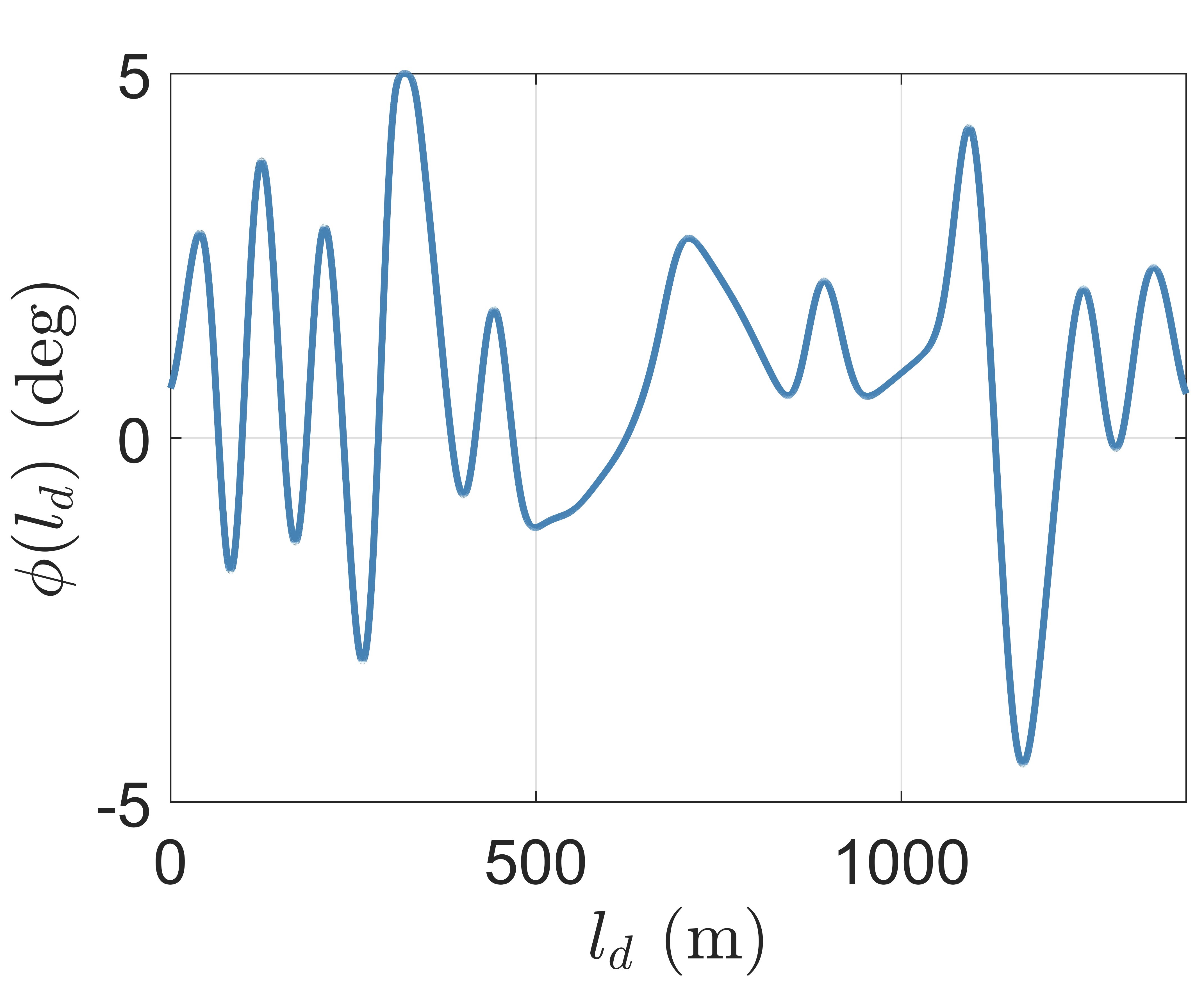}%
    }
    \caption{Setup for the heterogeneous-vehicle evaluation.}
    \label{fig:physical_setup}
\end{figure}

Figure~\ref{fig:physical_setup} shows the desired path, its curvature, the
longitudinal-speed profile, and the road bank angle. The path is $1390$~m long with a minimum radius of $25$~m. Speed follows the curvature magnitude: 23 m/s on
straight segments, decreasing to 6 m/s as the
curvature magnitude increases. Every vehicle is
assigned the same profile $v_x(l_d)$. The bank angle $\phi(l_d)$ is proportional to curvature up to $5^\circ$ and signed so the road banks into each turn. It enters the time-domain lateral dynamics through the gravity term
$g\sin\phi(l_d)$ and is unknown to all controllers. No perturbation is imposed directly on the reduced
arc-length dynamics. Instead, $w_i$ in
\eqref{eq:perturbed_arc_generic} represents the residual
effects of road bank, parameter mismatch, and the
modeling approximations.

Heterogeneity and uncertainty enter separately. Each vehicle is first
assigned a baseline parameter set $p_{i,0}=p_{\rm
ref}(1+\bar\Gamma_p\zeta_{i,p})$ with $\zeta_{i,p}\sim\mathcal U[-1,1]$, for
every $p\in\{m,I_z,a,b,C_f,C_r\}$, using
the reference values and heterogeneity bounds $\bar\Gamma_p$ of
Table~\ref{tab:physical_parameters}. These baseline parameters differ among vehicles,
and vehicle $i$'s controller uses its assigned
values $p_{i,0}$.
Parametric uncertainty is then applied on top, as $p_i^{\rm
true}=p_{i,0}(1+\bar\Delta\eta_{i,p})$ with $\eta_{i,p}\sim\mathcal U[-1,1]$
independent of $\zeta_{i,p}$, for $p\in\{m,I_z,C_f,C_r\}$: the plant uses
$p_i^{\rm true}$ while the controller retains $p_{i,0}$. The geometric
parameters $a$ and $b$ are heterogeneous across the fleet but are not
perturbed.

\begin{table}[!htb]
\centering
\footnotesize
\caption{Reference vehicle parameters and baseline heterogeneity bounds}
\renewcommand{\arraystretch}{1.6}
\begin{tabular}{ccc}
\hline
\hline
\textbf{Parameter} & \textbf{Reference value} & \textbf{Heterogeneity bound} \\
\hline
$m_{\rm ref}$     & $1896~{\rm kg}$               & $35\%$ \\
$I_{z,\rm ref}$   & $3803~{\rm kg\,m^2}$          & $35\%$ \\
$a_{\rm ref}$     & $1.2682~{\rm m}$               & $10\%$ \\
$b_{\rm ref}$     & $1.5818~{\rm m}$               & $10\%$ \\
$C_{f,\rm ref}$   & $4.00\times10^5~{\rm N/rad}$  & $35\%$ \\
$C_{r,\rm ref}$   & $3.819\times10^5~{\rm N/rad}$ & $35\%$ \\
\hline
\hline
\end{tabular}
\label{tab:physical_parameters}
\end{table}

At $\bar\Delta=0$, the vehicles remain heterogeneous,
but each controller uses its vehicle's true parameter
values. The common road-bank contribution to the
spatial perturbation is approximately
$g\sin\phi(l_d)/v_x^2(l_d)$.
Increasing $\bar\Delta$ introduces parameter mismatch,
which contributes vehicle-specific residuals. Two studies are
reported, the first at the fixed level $\bar\Delta=0.10$ over a
$100$-vehicle platoon, the second sweeping $\bar\Delta$ from $0$ to $0.75$
over $N=20$. All random variables are sampled once, and the
same realization is used for all three configurations
and throughout the uncertainty sweep. Thus,
increasing $\bar\Delta$ scales each vehicle's
parameter mismatch without resampling it. The initial conditions satisfy
$|e_{lat,i}^{des}(0)|\le0.05$~m
and $|(e_{lat,i}^{des})'(0)|\le0.01$.

\begin{figure}[h]
    \centering
    \subfloat[PT]{%
        \includegraphics[width=0.48\columnwidth]
        {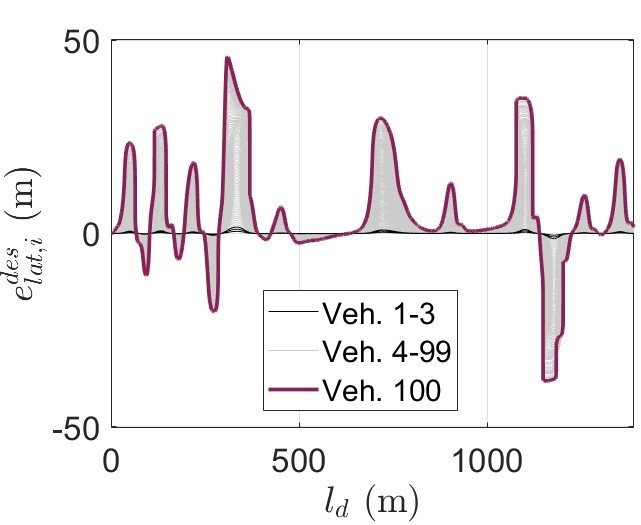}%
    }
    \hfil
    \subfloat[DT]{%
        \includegraphics[width=0.48\columnwidth]
        {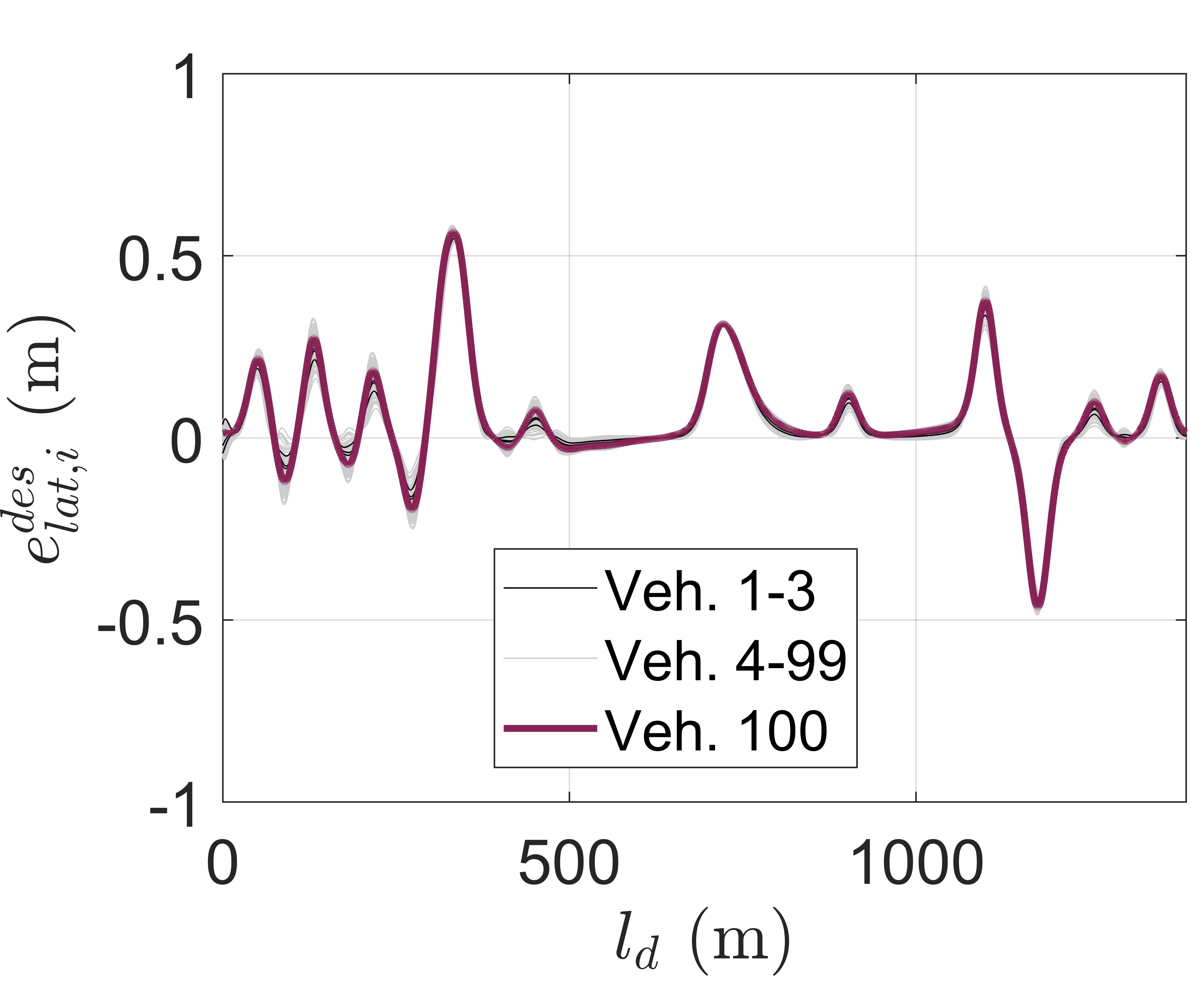}%
    }

    \subfloat[LFP]{%
        \includegraphics[width=0.48\columnwidth]
        {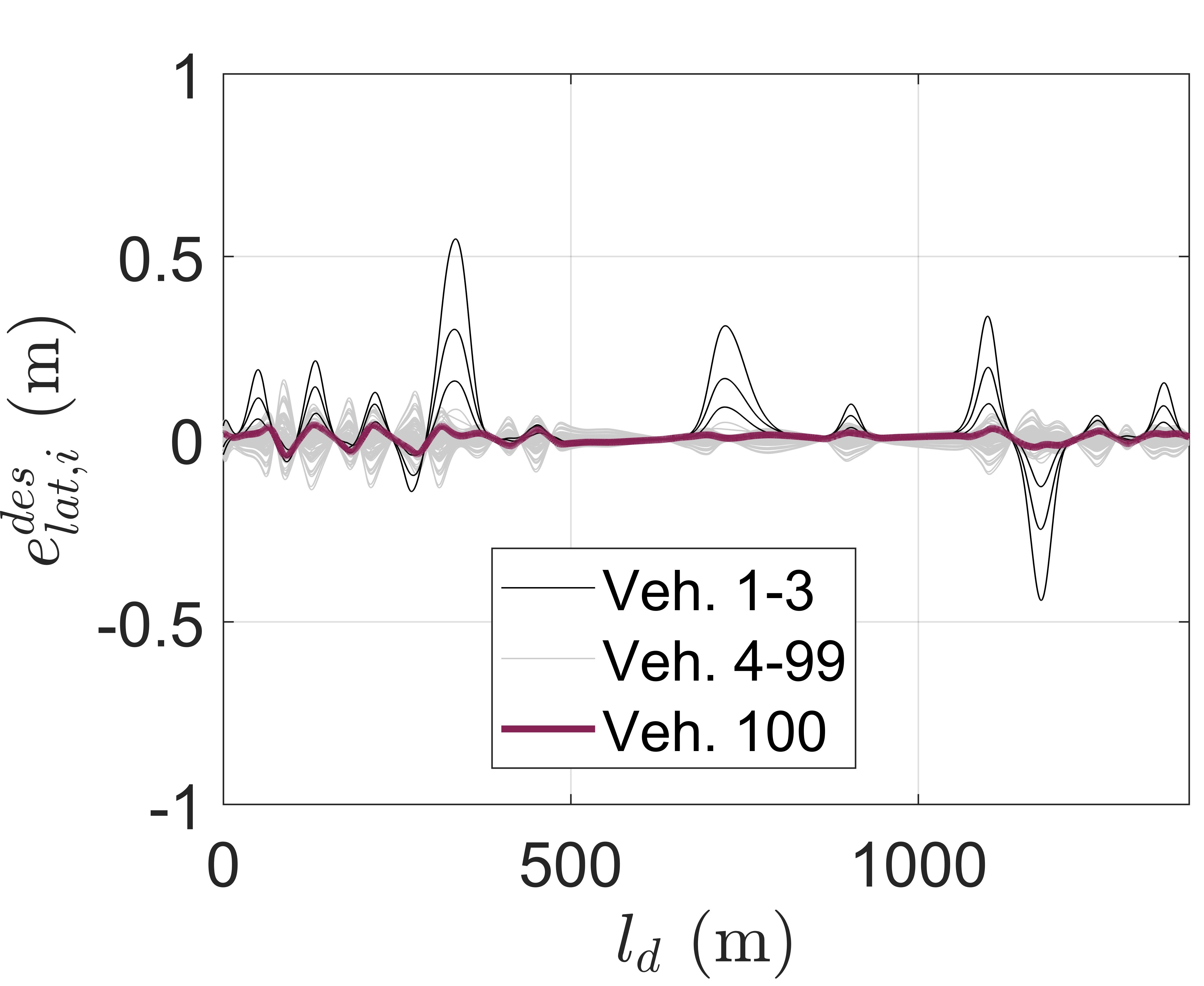}%
    }
    \caption{Desired-path lateral errors of $100$ heterogeneous
    vehicles at $\bar\Delta=0.10$. Panels (b) and (c) share
    a vertical scale; (a) uses a different scale.}
    \label{fig:physical_population}
\end{figure}

Figure~\ref{fig:physical_population} shows the desired-path lateral errors of
all $100$ vehicles at $\bar\Delta=0.10$, one panel per configuration, and
separates the same three behaviors on the full model. Under PT the
peak error grows with the vehicle index, from $0.548$~m at the leader to
$45.8$~m at vehicle $100$. Under DT, the vehicle errors remain near the leader's
level, with peak errors between $0.54$ and $0.57$~m
across the platoon and no systematic downstream
attenuation. Under LFP the
peak error falls over the first few vehicles and then settles, with vehicle
$100$ at $0.0466$~m, an order of magnitude below the DT level.

\begin{figure}[h]
    \centering
    \subfloat[$\mathcal L_2$ norm]{%
        \includegraphics[width=0.48\columnwidth]
        {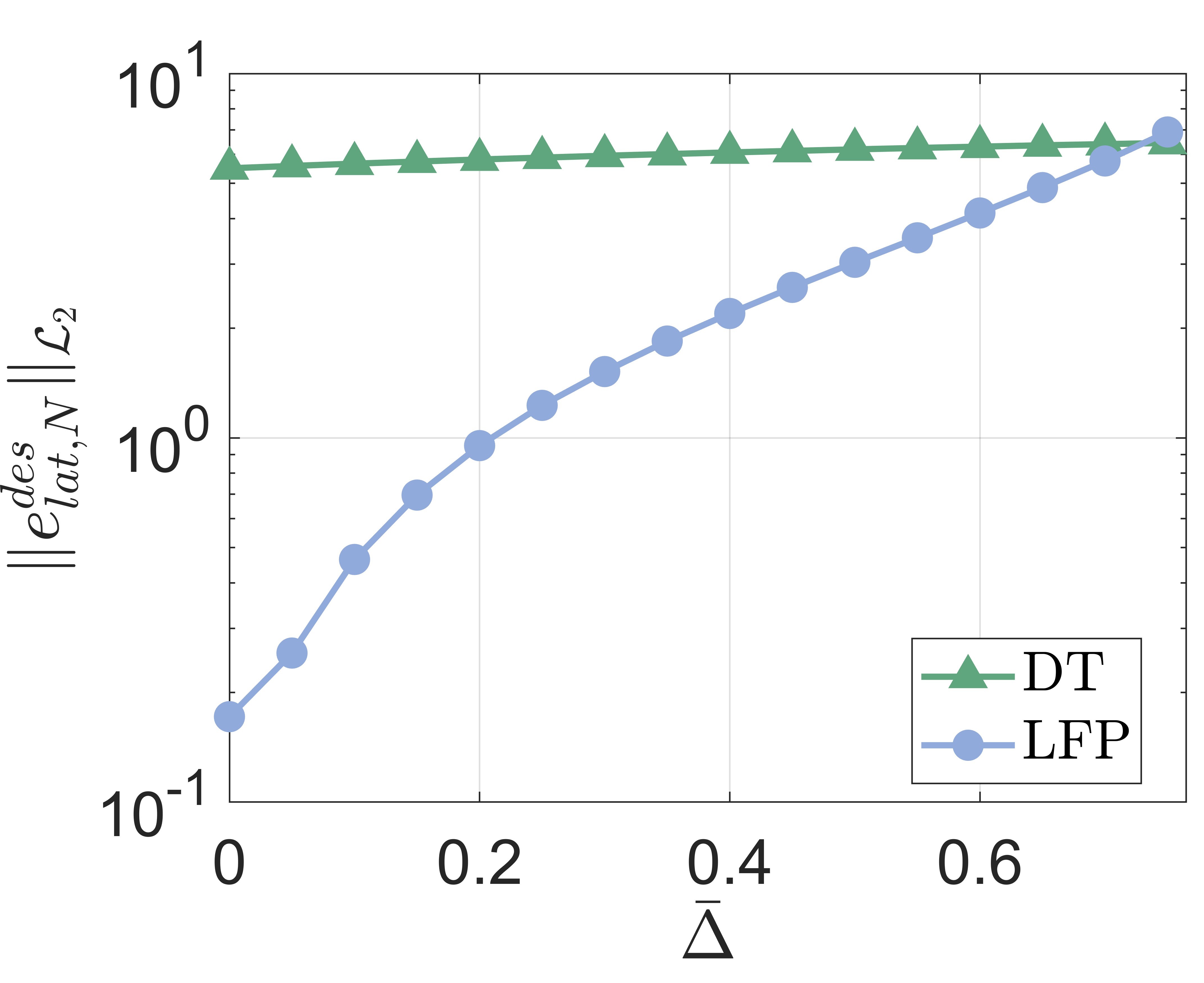}%
    }
    \hfil
    \subfloat[$\mathcal L_\infty$ norm]{%
        \includegraphics[width=0.48\columnwidth]
        {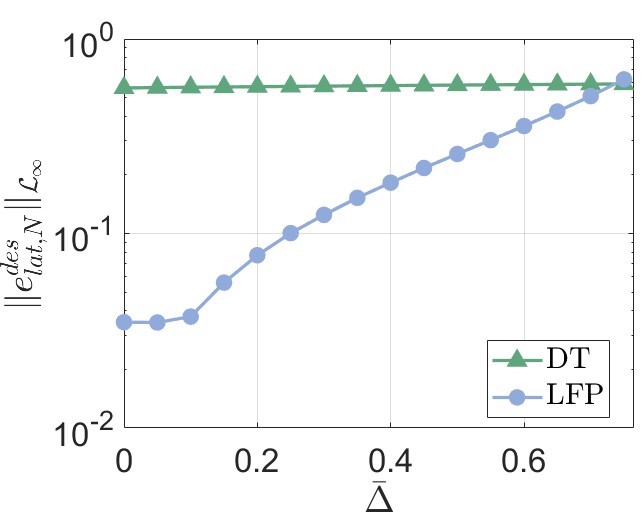}%
    }
    \caption{Last-vehicle lateral error norms versus parameter
    uncertainty level $\bar\Delta$ for $N=20$.}
    \label{fig:physical_sweep}
\end{figure}

Figure~\ref{fig:physical_sweep} shows the last-vehicle
norms against $\bar\Delta$ for the fixed parameter
realization used in this study. The LFP norms increase
more rapidly than the DT norms, narrowing the
performance gap as uncertainty increases.
At $\bar\Delta=0.5$, LFP remains approximately a factor
of two below DT in both norms. The curves meet near $\bar\Delta=0.73$, corresponding
to parameter deviations of up to $73\%$ from the values
used by the controllers.

To characterize perturbation repeatability in the
vehicle-model study, we evaluate
$w_i=(e_{lat,i}^{des})''-\mu_i$ under DT.
Define the common component as the platoon mean,
$w_c=N^{-1}\sum_{i=1}^N w_i$, and let
$\tilde w_i=w_i-w_c$. Since $\sum_i\tilde w_i=0$,
\begin{equation}
\sum_{i=1}^N\|w_i\|_{\mathcal L_2}^2
=
N\|w_c\|_{\mathcal L_2}^2
+\sum_{i=1}^N\|\tilde w_i\|_{\mathcal L_2}^2.
\end{equation}
We quantify the vehicle-specific share by
\begin{equation}
S=
\left(
\frac{\sum_{i=1}^N\|\tilde w_i\|_{\mathcal L_2}^2}
     {\sum_{i=1}^N\|w_i\|_{\mathcal L_2}^2}
\right)^{1/2}.
\end{equation}
For nonzero total perturbation energy, $S\in[0,1]$.
This provides a measure of the residual share under DT,
analogous to the prescribed ratio $\beta/\bar w$ in
the reduced-model study.

Across the sweep, $S$ is near zero at $\bar\Delta=0$,
rises to $0.60$ at $\bar\Delta=0.5$, and reaches $0.87$
at the end. Thus, increasing parameter uncertainty
shifts the perturbation from predominantly common
to predominantly vehicle-specific.

     \section{Conclusion}
\label{sec:chapter4_summary}

This paper studied robust lateral string stability and
error attenuation for heterogeneous vehicle platoons
under varying longitudinal motion. An arc-length-based
feedback linearization formulation enabled platoon-level
analysis with vehicle-specific dynamics compensated
locally and perturbations retained explicitly.
Within this framework, we introduced a robust
$\mathcal L_2$ lateral string stability notion and
established a sufficient condition for one-predecessor
LTI interconnections. The analysis further provided
a platoon-length-uniform $\mathcal L_\infty$ lateral
error bound.

The results distinguish the benefits of desired-path
availability from those of predecessor-based learning.
Under the PT formulation, common local perturbations
can produce lateral errors that accumulate along
the platoon.
Sharing the desired path removes this accumulation
in the DT baseline, while the proposed LFP strategy
uses predecessor information to achieve additional
error attenuation. An explicit learning-gain choice
guarantees robust $\mathcal L_2$ lateral string
stability and yields geometric attenuation under
common perturbations and zero initial conditions.
When perturbations differ among vehicles, the derived
error envelopes quantify how adjacent-vehicle
perturbation differences limit this attenuation.
Reduced-model simulations validated the theoretical
bounds and attenuation envelopes, while simulations
with heterogeneous vehicle dynamics demonstrated
the three configurations' behavior under varying
longitudinal speed, parameter uncertainty, and
road bank. Together, the results establish how intervehicle
information can support bounded lateral errors
and quantify the role of perturbation repeatability
in downstream error attenuation through LFP.
 
\appendix
\subsection{State-Space Realization of the V2V Error Dynamics}
\label{app:v2v_realization}
To verify that the V2V error dynamics belong to the class of LTI
interconnections considered in
\eqref{eq:general_lti_leader}--\eqref{eq:general_lti_follower}, the
realization must reproduce both maps of \eqref{eq:v2v_transfer_recursion}:
the coupling to the predecessor must enter only through its performance
output $y_{i-1}=e_{lat,i-1}^{des}$, with transfer function $H$, and the
local input must enter through $T_{y\psi}=1/D$. The follower recursion \eqref{eq:lfp_recursion}
also involves the first and second arc-length derivatives of that output, so
the realization below absorbs the first derivative into the state and
the second into the feedthrough. Separating the feedthrough
$D_p=H(\infty)$ from \eqref{eq:v2v_H} gives
\begin{equation}
\label{eq:v2v_realization_split}
H(s)
=
D_p
+
\frac{\beta_1s+\beta_0}{s^2+k_2s+k_1},
\end{equation}
where
\begin{align}
\label{eq:v2v_realization_coefficients}
D_p
&:=
1+k_{LDD},
\qquad
\beta_1
:=
k_{LD}-k_2k_{LDD},
\notag \\
\beta_0
&:=
k_{LP}-k_1k_{LDD}.
\end{align}
The strictly proper part of \eqref{eq:v2v_realization_split} has the
observable canonical realization
\begin{equation}
\label{eq:v2v_realization_matrices}
A=
\begin{bmatrix}
-k_2 & 1\\
-k_1 & 0
\end{bmatrix},
~
B_p=
\begin{bmatrix}
\beta_1\\
\beta_0
\end{bmatrix},
~
B_\psi=
\begin{bmatrix}
0\\
1
\end{bmatrix},
~
C_y=
\begin{bmatrix}
1 & 0
\end{bmatrix},
\end{equation}
with $D_p$ as the feedthrough, so that $D_p+C_y(sI-A)^{-1}B_p=H(s)$; the
input matrix $B_\psi$ realizes the local channel, since
$C_y(sI-A)^{-1}B_\psi=1/D(s)=T_{y\psi}(s)$. The realization therefore
reproduces both maps of \eqref{eq:v2v_transfer_recursion}.

The state follows from the realization itself. The output equation
$e_{lat,i}^{des}=C_yx_i+D_pe_{lat,i-1}^{des}$ with $C_y=[1\ \ 0]$ gives
$x_{i,1}=\zeta_i$, where $\zeta_i:=e_{lat,i}^{des}-D_pe_{lat,i-1}^{des}$, and
solving
the first row of $x_i'=Ax_i+B_pe_{lat,i-1}^{des}+B_\psi\psi_i$ for $x_{i,2}$
gives
\begin{equation}
\label{eq:v2v_realization_state}
x_i
=
\begin{bmatrix}
\zeta_i\\
\zeta_i'+k_2\zeta_i-\beta_1e_{lat,i-1}^{des}
\end{bmatrix},
\quad i=2,\ldots,N,
\end{equation}
and, for the lead vehicle, $\zeta_1:=e_{lat,1}^{des}$ with
$x_1=[\zeta_1,\ \zeta_1'+k_2\zeta_1]^T$.
The realization state $x_i$ need not coincide with the
error state $x_{e,i}$. Its dynamics and the relationship
between $x_i(0)$ and $x_{e,i}(0)$ are established below.

With \eqref{eq:v2v_realization_state}, \eqref{eq:v2v_leader} and
\eqref{eq:lfp_recursion} can be written as
\begin{align}
x_1'
&=
Ax_1+B_\psi w_1, \qquad
e_{lat,1}^{des}=
C_yx_1, 
\\
x_i'
&=
Ax_i+B_pe_{lat,i-1}^{des}+B_\psi\left(w_i-w_{i-1}\right), \notag \\
e_{lat,i}^{des}
&=
C_yx_i+D_pe_{lat,i-1}^{des},
\qquad i=2,\ldots,N,
\end{align}
so that the local inputs in
\eqref{eq:general_lti_leader}--\eqref{eq:general_lti_follower} are
$\psi_1=w_1$ and $\psi_i=w_i-w_{i-1}$ for $i=2,\ldots,N$.
The characteristic polynomial of $A$ is $s^2+k_2s+k_1$, so $A$ is Hurwitz
for $k_1>0$ and $k_2>0$. Under the gains \eqref{eq:alpha_gains}, $H\equiv1-\alpha$ by
\eqref{eq:alpha_H}, so its strictly proper part in
\eqref{eq:v2v_realization_split} vanishes: $D_p=1-\alpha$,
$\beta_1=\beta_0=0$, and hence $B_p=0$. The coupling is the feedthrough
alone, $e_{lat,i}^{des}=C_yx_i+\left(1-\alpha\right)e_{lat,i-1}^{des}$.

To relate the initial realization states to the initial
error states, recall from
Section~\ref{sec:robust_lateral_string_stability} that
\begin{equation}
    x_{e,i}(0)
    =
    \begin{bmatrix}
        e_{lat,i}^{des}(0)\\
        \big(e_{lat,i}^{des}\big)'(0)
    \end{bmatrix}.
\end{equation}
Since $|e_{lat,i}^{des}(0)|$ and $|(e_{lat,i}^{des})'(0)|$ are each bounded by
$\|x_{e,i}(0)\|_2$, bounding the two components of
\eqref{eq:v2v_realization_state} separately and using
$\|x_i(0)\|_2\le|x_{i,1}(0)|+|x_{i,2}(0)|$ gives, for every follower,
\begin{align}
\|x_i(0)\|_2
&\le
\left(1+|D_p|\right)
\max_{1\le j\le N}\|x_{e,j}(0)\|_2
\notag\\
&\quad+
\left[
\left(1+|D_p|\right)\left(1+k_2\right)+|\beta_1|
\right]
\max_{1\le j\le N}\|x_{e,j}(0)\|_2
\notag\\
&=
\left[
\left(1+|D_p|\right)\left(2+k_2\right)+|\beta_1|
\right]
\max_{1\le j\le N}\|x_{e,j}(0)\|_2.
\end{align}
The same bound holds for the leader, for which $D_p$ and $\beta_1$ are
absent. Hence,
{\small
\begin{equation}
\max_{1\le i\le N}\|x_i(0)\|_2
\le
\left[(1+|D_p|)(2+k_2)+|\beta_1|\right]
\max_{1\le j\le N}\|x_{e,j}(0)\|_2.
\end{equation}
}
Therefore, the initial realization states $x_i(0)$ are
uniformly bounded in norm by the initial error states
$x_{e,i}(0)$, with a constant independent of the
platoon length.

\bibliographystyle{IEEEtran}
\bibliography{IEEEabrv,root}

\end{document}